%% file: main.tex
\documentclass[conference]{IEEEtran}
\pdfoutput=1 

\usepackage[noadjust]{cite} 
\usepackage[acronym]{glossaries}
\usepackage{multirow}
\usepackage{amssymb}
\usepackage{mathtools}
\DeclarePairedDelimiter\ceil{\lceil}{\rceil}
\DeclarePairedDelimiter\floor{\lfloor}{\rfloor}
\usepackage{grffile}
\usepackage{booktabs} 
\usepackage{subfig}
\usepackage{flushend}
\usepackage{times}
\usepackage{textcomp}
\usepackage{tabulary}
\usepackage{color}
\usepackage{colortbl}

\newcommand{\snip}[1]{}

\usepackage[font={small,it}]{caption}

\begin{document}
%
\title{Low-memory GEMM-based convolution
  algorithms for deep neural networks}

\author{\IEEEauthorblockN{Andrew Anderson\IEEEauthorrefmark{1} \hspace{1cm} Aravind Vasudevan\IEEEauthorrefmark{1}\IEEEauthorrefmark{2} \hspace{1cm}Cormac Keane\IEEEauthorrefmark{1} \hspace{1cm}David Gregg\IEEEauthorrefmark{1}}
  \IEEEauthorblockA{\IEEEauthorrefmark{1}Lero, School of Computer Science \& Statistics, Trinity College Dublin}
  \IEEEauthorblockA{\IEEEauthorrefmark{2}Synopsys Inc, Dublin\\
  \{aanderso, aravin, ckeane4, dgregg\}@tcd.ie}
}
\input{definitions}
\maketitle

\glsunset{i2c}
\begin{abstract}
  
  Deep neural networks (DNNs) require very large
  amounts of computation both for training and for inference when
  deployed in the field. A common approach to implementing DNNs is to
  recast the most computationally expensive operations as general
  matrix multiplication (GEMM). However, as we demonstrate in this
  paper, there are a great many different ways to express DNN
  convolution operations using GEMM. Although different approaches all
  perform the same number of operations, the size of temorary data
  structures differs significantly. Convolution of an input matrix
  with dimensions $C \times H \times W$, requires $O(K^2CHW)$
  additional space using the classical \textit{im2col} approach. More
  recently memory-efficient approaches requiring just $O(KCHW)$
  auxiliary space have been proposed.

  We present two novel GEMM-based algorithms that require just
  $O(MHW)$ and $O(KW)$ additional space respectively, where $M$ is the
  number of channels in the result of the convolution. These
  algorithms dramatically reduce the space overhead of DNN
  convolution, making it much more suitable for memory-limited
  embedded systems. Experimental evaluation shows that our low-memory
  algorithms are just as fast as the best patch-building approaches
  despite requiring just a fraction of the amount of additional
  memory. Our low-memory algorithms have excellent data locality which
  gives them a further edge over patch-building algorithms when
  multiple cores are used. As a result, our low memory algorithms
  often outperform the best patch-building algorithms using multiple
  threads.

\end{abstract}

\IEEEpeerreviewmaketitle

\input{introduction}

\input{background}

\input{methodology}
\input{results}

\input{related-work}



\section{Conclusion} Multi-channel multi-kernel convolution is the most
computationally expensive operation in \glspl{dnn}. Maximal
exploitation of processor resources for \gls{mcmk} requires a deep
understanding of the micro-architecture. Careful design of data
blocking strategies to exploit caches, on-chip memories and register
locality are needed, along with careful consideration of data movement
and its interaction with SIMD/SIMT parallelism. Each new processor has
different performance characteristics, requiring careful tuning of the
code each time it is brought to a new target.

There are significant advantages in implementing \gls{mcmk}
convolution using existing carefully tuned \glsfirst{gemm} libraries.
However, the most widely-used approach, \gls{i2c} has a large memory
footprint because it expands the input image to a much larger patch
matrix. This space expansion is quadratic in the order, $K$, of the 2D
convolution being performed. This is problematic for
memory-constrained systems such as embedded object detection and
recognition systems. Additionally, the data redundancy resulting from
\gls{i2c} reduces data locality and increases memory traffic.

We propose two new approaches for implementing MCMK convolution using
existing parallel \gls{gemm} libraries that require much less memory.
These algorithms dramatically reduce the space overhead of DNN
convolution, making it much more suitable for memory-limited embedded
systems. These algorithms replace the single GEMM call of the $K^2$
patch-building approaches with an accumulation of the result of $K^2$
different $1 \times 1$ convolutions. The result is a dramatic reduction
in memory requirements and improved data locality.

Experimental evaluation shows that our low-memory algorithms
are just as fast as the best patch-building approaches despite
requiring just a fraction of the amount of additional memory. Our
low-memory algorithms have excellent data locality which gives them a
further edge over patch-building algorithms when multiple cores are
used. As a result, our low memory algorithms often outperform the best
patch-building algorithms using multiple threads.

\section*{Acknowledgment}

\setlength{\lineskip}{0.5em}
\noindent This work was completed when author Aravind Vasudevan was a
postdoctoral researcher at Trinity College Dublin. He is now at
Synopsys Inc. This work was supported by Science Foundation Ireland
grant 12/IA/1381; the European Union’s Horizon 2020 research and
innovation programme under grant agreement No 732204 (Bonseyes); and
in part by Science Foundation Ireland grant 13/RC/2094 to Lero --- the
Irish Software Research Centre (www.lero.ie).

%
%

\bibliographystyle{IEEEtran}
\bibliography{references}
\input{resultstable}
\end{document}

%% file: definitions.tex
\newcommand{\aravind}[1]{*** AV : {\color{red} #1 }***}
\newcommand{\andrew}[1]{*** AA : {\color{blue} #1 }*** }
\newcommand{\david}[1]{*** DG : {\color{green} #1 }***}
\newcommand{\needsmassaging}[1]{Needs some massaging --- {\color{blue} #1 }}

\newcommand*\concat{\mathbin{\|}}

\newacronym{dnn}{DNN}{Deep Neural Network}
\newacronym{cnn}{CNN}{Convolutional Neural Network}
\newacronym{mec}{MEC}{Memory-Efficient Convolution}

\newacronym{scsk}{SCSK}{Single Channel Single Kernel}
\newacronym{mcsk}{MCSK}{Multiple Channel Single Kernel}
\newacronym{mcmk}{MCMK}{Multiple Channel Multiple Kernel}

\newacronym{gemv}{GEMV}{General Matrix Vector multiplication}
\newacronym{gemm}{GEMM}{General Matrix Multiplication}

\newcommand{\ifm}{\mathcal{I}}			
\newcommand{\ofm}{\mathcal{O}}		 
\newcommand{\kernel}{\mathcal{K}}

\newcommand{\iw}{W}
\newcommand{\ih}{H}
\newcommand{\ic}{C}

\newcommand{\kw}{K}
\newcommand{\kh}{K}
\newcommand{\kc}{C}
\newcommand{\kn}{M}

\newcommand{\ow}{P}
\newcommand{\oh}{Q}
\newcommand{\oc}{\kn}

\newacronym{i2c}{$im2col$}{im2col}
\newacronym{i2r}{$im2row$}{im2row}


\newacronym{i2c-scan}{$im2col-scan$}{}
\newacronym{i2c-short}{$im2col-copy-short$}{}
\newacronym{i2c-long}{$im2col-copy-long$}{}
\newacronym{i2c-self}{$im2col-copy-self$}{}
\newacronym{i2r-scan}{$im2row-scan$}{}
\newacronym{i2r-short}{$im2row-copy-short$}{}
\newacronym{i2r-long}{$im2row-copy-short$}{}


\newacronym{googlenet}{GoogLeNet}{GoogLeNet}
\newacronym{alexnet}{AlexNet}{AlexNet}
\newacronym{vgg}{VGG-16}{VGG-16}
\glsunset{googlenet}
\glsunset{alexnet}
\glsunset{vgg}

\newacronym{k2r}{$kn2row$}{Kernel to Row} 
\newacronym{k2c}{$kn2col$}{Kernel to Column} 

\newacronym{k2r-as}{$kn2row-as$}{Accumulating Kernel to Row} 
\newacronym{k2r-aa}{$kn2row-aa$}{Hole Punching Accumulating Kernel to Row} 
\newacronym{k2r-aa-pp}{$kn2row-aa-pp$}{Accumulating Kernel to Row with Post-pass} 
\newacronym{k2r-aa-hp}{$kn2row-aa-hp$}{Hole Punching Accumulating Kernel to Row} 
\newacronym{k2r-an}{$kn2row-acc-noshift$}{Accumulating \gls{k2r} without Output Shifting} 
%
\newacronym{k2c-as}{$kn2col-as$}{Accumulating Kernel to Column} 

%% file: introduction.tex
\section{Motivation}
\label{sec:motivation}

Deep neural networks are among the most successful techniques for
processing image, video, sound and other data arising from real-world
sensors. DNNs require very large amounts of computation which
challenge the resource of all but the most powerful machines.
However, DNNs are also most useful when deployed in embedded devices
that interact directly with the surrounding world. Thus there is a
tension between the resources needed by deep neural networks, and the
kind of embedded devices that can make best use of deep learning
technology by applying it to live data from embedded sensors.

DNNs consist of an acyclic directed graph of ``layers'' that receive
raw input data, and commonly output a classification of the data.
Data flows along edges between layers. Each layer processes its input
data, and produces output data on its outgoing edges.  Several
different types of layers are used to implement DNNs, such as
activation layers, pooling layers, convolution layers, and
fully-connected layers. In the best-known DNNs a great majority of the
execution time is spent in the convolution layers.

There are many ways that each layer can be implemented. For example,
early DNN libraries simply implemented each layer with a loop nest.
However, Yangqing Jia discovered that the convolution layers could be
implemented more efficiently by restructuring the data and calling a
standard matrix multiplication routine. Most machines already have a
fast implementation of the Basic Linear Algebra Subprograms (BLAS),
which includes a general matrix multiplication (GEMM) routine.  BLAS
libraries are carefully hand-coded, and may include code generation
and auto-tuning systems to maximize performance. Implementing
convolution using GEMM allows the developer to exploit these
highly-tuned routines.

DNN layers can be implemented in other ways. For example, many layers
are implemented as carefully coded loop nests that provide a direct
implementation of the layer \cite{jeffers2016intel}. Another common approach
is to convert the data to another format, such as the Fourier domain,
where convolution can be computed
efficiently \cite{Highlander:2016}. Several authors have also proposed
domain-specific code generators, that can generate many variants of
the code for a problem and use auto-tuning to select a fast
version \cite{Truong:2016}.

We study a variety of
different approaches to implementing DNN convolution using the BLAS
GEMM routine. By rewriting the convolution as different variants of
GEMM calls with different data layouts, we can trade off execution
time, memory requirements and parallelism within a framework that is
suitable for automation.
Our main contributions are as follows:
\begin{itemize}
\item We provide the first systematic study of the design space of DNN
  convolution implementation using GEMM.
\item We note that the most popular existing patch-building DNN convolution algorithms require
 $O(CHWK^2)$ additional temporary space. Even the most
 memory-efficient GEMM-based DNN convolution algoirithm needed
 $O(CHWK)$ additional space.
\item We propose two novel GEMM-based algorithms for
  DNN convolution based on accumulating the results of smaller
  convolutions. The two algorithms require only $O(MHW)$ and $O(KW)$
  additional space respectively.
\item We evaluate a very large range of GEMM-based convolution algorithms
and variants on Intel Core i7 and ARM Cortex A57 processors. We find
that our low-memory DNN convolution algorithms can be just as fast as
the best patch-building algorithms despite needing only a fraction of
the space.
\item In addition we find that when multiple cores are used to perform
the convolution, the low memory requirements of our algorithms
improves locality to the point that they are often faster than
equivalent patch-building algorithms.
\end{itemize}


%% file: background.tex

%

\section{DNN convolution}
\label{sec:back}



DNNs consist of a directed graph of standard \textit{layers} which
operate on data on incoming edges and place results on outgoing
edges. In many DNNs, the majority of execution time is occupied by
\textit{convolution layers}. A convolutional layer takes operates
on two source tensors and produces another tensor as output. In C code, these
tensors might be repesented as:

\begin{verbatim}
float input[C][H][W];
float kernels[M][C][K][K];
float output[M][H][W];
\end{verbatim}


A convolution layer consists of four components: 1) a 3D tensor $\ifm \in
\mathbb{R}^{\ih\times\iw\times\ic}$ which acts as as the input to the
convolution layer 2) a set of ``kernels'' or ``filters'' represented by a 4D
tensor $\kernel
\in \mathbb{R}^{\kn\times\kc\times\kh\times\kw}$ 3) a bias term per filter 4)
and a 3D output tensor. \glsfirst{mcmk} is typically constructed as the
concatenation of the output of $\kn$ \glsfirst{mcsk} as described by Vasudevan et al. \cite{Vasudevan:2017} is often represented as nested summations as,

\begin{figure}[h]
	\begin{minipage}{1\linewidth}
		\begin{footnotesize}
			\begin{verbatim}
			input[C][H][W];
			kernels[M][K][K][C];
			output[M][H][W];
			for h in 1 to H do
			  for w in 1 to W do
			    for o in 1 to M do
			      sum = 0;
			      for x in 1 to K do
			        for y in 1 to K do
			          for i in 1 to C do
			            sum += input[i][h+y][w+x]
			                   *kernels[o][x][y][i];
			      output[o][w][h] = sum;
			\end{verbatim}
		\end{footnotesize}
	\end{minipage}
	\caption{Simplified code for 2D multi-channel convolution with a single multi-channel input and multiple multi-channel convolution kernels. Note that special treatment of edge boundaries is not shown in this code.}
	\label{fig:covCode}
\end{figure}

\section{Convolution with $O(K^2CHW)$ patch matrix}
\label{sec:im2c}
An attractive approach to implementing DNN convolution is to recast it as
matrix multiplication and use highly optimized routines such as general
matrix multiplication (GEMM) to find the solution.
The \gls{i2c} approach~\cite{chellapilla2006high,tsai2016performance,yanai2016efficient,gu2016opencl} has been well studied for transforming the \gls{mcmk} problem into a \gls{gemm} problem. Consider an input $\ifm \in \mathbb{R}^{\ih\times\iw\times\ic}$ and $\kn$ kernels $\kernel\in\mathbb{R}^{\kn\times\kh\times\kw\times\kc}$. From the input $\ifm$ we construct a new \emph{input-patch-matrix} $\hat{\ifm}$, by copying \emph{patches} out of the input and unrolling them into columns of this intermediate matrix. These patches are formed in the shape of the kernel (i.e. $\kh\times\kw\times\kc$) at every location in the input where the kernel is to be applied.


Once the input-patch-matrix $\hat{\ifm}$ is formed, we construct the kernel-patch-matrix $\hat{\kernel}$ by unrolling each of the $\kn$ kernels of the shape $\kh\times\kw\times\kc$ into a row of $\hat{\kernel}$. Note that this step can be avoided if the kernels are stored in this format to begin with (innermost dimension is the channel which forces the values along a channel to be contiguous). Then we simply perform a \gls{gemm} of $\hat{\kernel}$ and $\hat{\ifm}$ to get the output $\hat{\ofm}\in\mathbb{R}^{\ih\times\iw\times\kn}$ as shown in the figure.


\subsection{Matrix layouts}
The inputs and outputs of convolution are multidimensional tensors, so
there are a great many possible orderings of the dimensions. Many DNN
implementations use a single canonical tensor layout. For example,
Caffe~\cite{jia2014caffe} lays out the dimensions of input data tensors in the
order $C \times H \times W$ (or more briefly $CHW$), and builds patch
matrices with dimensions $CKKHW$. When invoking the GEMM operation the
patch matrix is interpreted as a 2D matrix of dimensions $CKK \times
HW$.

Building the patch matrix is a simple operation. But any dimension
ordering might be used for the input or patch matrix, so there are
many variants. One possible method for implementing \gls{i2c} is by 
creation of patches that each occupy a column of the patch matrix. It
is easy to imagine another variant where each patch instead occupies a
row of the resulting matrix. Vasudevan et al. \cite{Vasudevan:2017} refer
to this variant as \gls{i2r}.

An additional complication when describing these variants is that
there are two competing layouts of rows and columns in memory.
Fortran uses a \textit{column-major} format, where matrix elements
that are adjacent in memory are interpreted as being in the same row
of the matrix. C/C++ use \textit{row-major} format, where elements
from the same row are adjacent in memory. GEMM is originally a Fortran
routine, and therefore assumes a column-oriented layout, but many DNN
implementations are in C/C++, which may result in confusion about
array formats in memory. For this reason we often refer to patch
matrices as being \textit{patch-major} or \textit{KKC-major} meaning
that elements of the same patch are adjacent in memory. Another
possible layout is \textit{HW-major} or \textit{patch-minor}, where
elements of the non-patch dimension of the patch matrix are adjacent
in memory.

\begin{figure}
	\centering
\begin{verbatim}
float input[H][W][C];
float patches[H][W][K][K][C];
for ( h = 0; h < H; h++ )
  for ( w = 0; w < W; w++ )
    for ( kh = -K/2; kh < K/2; kh++ )
      for ( kw = -K/2; kw < K/2; kw++ )
        for ( c = 0; c < C; c++ )
          patches[h][w][kh][kw][c]
                 = input[h+kh][w+kw][c];
\end{verbatim}
\vspace{-0.25cm}
\caption{Building the $KKC$-major patch matrix}
\label{fig:im2row-code}
\end{figure}

Figure \ref{fig:im2row-code} shows a loop nest for constructing a
patch matrix from the input. The patch matrix has
dimentions \textit{HWKKC}, which allows us to consider as a five
dimensional $H \times W \times \times K \times K \times C$ matrix, or
a two-dimensional $HW \times KKC$ matrix. The C/C++ code in
Figure \ref{fig:im2row-code} assumes a row-major layout, which implies
a \textit{patch-major} or \textit{KKC-major} format.

A key step of building the patch matrix is gathering elements of
the \texttt{input} matrix. As shown in Figure \ref{fig:im2row-code},
the major dimension of \texttt{input} is the $C$ dimension. The loop
nest in Figure \ref{fig:im2row-code} copies items from
the \texttt{input} to the patch matrix. The innermost loop copies
adjacent elements of \textit{C-major} \texttt{input} matrix
to adjacent elements of the patch matrix. The result is that
the code in Figure \ref{fig:im2row-code} has extremely good
spatial locality, and is therefore very fast in practice.

\subsection{Patch-minor layouts}
Other layouts can result in very different data locality. For example,
consider the case where the patch matrix is \textit{HW-major} rather
than \textit{patch-major}. Figure \ref{tab:2d-patch-matrix} shows an
example of such a matrix where, for simplicity we have assumed $C=1$.
If we were to construct the shaded patch column of
Figure \ref{tab:2d-patch-matrix} by gathering the corresponding shaded
patch elements of Figure \ref{tab:2d-input} in sequence, the spatial
locality would be extremely poor.

However, an alternative strategy results in much better spatial
locality. Note that the each row of the patch matrix contains a row of
the input. Therefore, if the input format is compatible, even a
non-\textit{patch-major} patch matrix can be constructed with
excellent spatial locality. Furthermore, even if the input does is not
suitable for directly copying rows to the patch matrix, it is
nonetheless possible to achieve almost as good spatial locality. Note
the successive rows of Figure \ref{tab:2d-patch-matrix} contain
repeated sequences of data. Once the first row containing that data
has been gathered, it is therefore possible to simply copy data from
one row of the patch matrix to the next.

This strategy of copying sequences of data from one row of a patch
matrix to the next is particularly important for strided DNN
convolution. A patch matrix for a convolution with stride $S$ has only
$HW/S$ patches as shown in Figure
\ref{tab:strided-patch-matrix}. Thus, the sequences of values across
the rows of a column-oriented patch matrix do not match the original
matrix layout. However, once the first row containing a particular set
of data has been collected, the data can be copied to later rows
containing the same sequence of values. To our knowledge we are the
first to exploit spatial locality between rows in this way, when
computing column-oriented patch matrices in row-major matrix layouts.

\begin{figure}[t!]
  \centering
\subfloat[Input matrix]{
  \label{tab:2d-input}
  \centering
\begin{tabular}{|c|c|c|c|c|}
\hline
\cellcolor[gray]{0.85}a0 & \cellcolor[gray]{0.85} a1 & \cellcolor[gray]{0.85} a2 &  a3 & a4 \\
 \cellcolor[gray]{0.85}b0 & \cellcolor[gray]{0.85}b1 & \cellcolor[gray]{0.85}b2 &b3 & b4 \\
\cellcolor[gray]{0.85}c0 & \cellcolor[gray]{0.85}c1 & \cellcolor[gray]{0.85}c2 & c3 & c4 \\
d0 & d1 & d2 & d3 & d4 \\
\hline
\end{tabular}
}

\subfloat[Patch matrix of $3 \times 3$ patches]{
  \label{tab:2d-patch-matrix}
  \centering
  \setlength\tabcolsep{5pt}
\begin{tabular}{|c|c|>{\columncolor[gray]{0.85}}c|c|c|c|c|}
 \hline
 0 &  0 & a0 & a1 & a2 & a3 & a4\\ 
 0 & a0 & a1 & a2 & a3 & 04 &  0\\
a0 & a1 & a2 & a3 & a4 &  0 &  0\\
 0 &  0 & b0 & b1 & b2 & b3 & b4\\
 0 & b0 & b1 & b2 & b3 & b4 &  0\\
b0 & b1 & b2 & b3 & b4 &  0 &  0\\
 0 &  0 & c0 & c1 & c2 & c3 & c4\\
 0 & c0 & c1 & c2 & c3 & c4 &  0\\
c0 & c1 & c2 & c3 & c4 &  0 & 0\\
 \hline
\end{tabular}
}
\subfloat[Stride 2 patch matrix]{
  \label{tab:strided-patch-matrix}
  \centering
  \setlength\tabcolsep{5pt}
\begin{tabular}{|c|>{\columncolor[gray]{0.85}}c|c|c|c|c|}
 \hline
 0 & a0 & a2 & a4\\ 
 0 & a1 & a3 &  0\\
a0 & a2 & a4 &  0\\
 0 & b0 & b2 & b4\\
 0 & b1 & b3 &  0\\
b0 & b2 & b4 &  0\\
 0 & c0 & c2 & c4\\
 0 & c1 & c3 &  0\\
c0 & c2 & c4 & 0\\
 \hline
\end{tabular}
}
\caption{Sub-parts of input matrix and column patch matrix, and strided patch matrix with patches of size $3 \times 3$. The shaded part is a corresponding patch in each matrix.
The outermost dimension, $C$, is not shown.}
\end{figure}

\snip{
\begin{table}[]
  \centering
    \setlength\tabcolsep{0.5pt}
\begin{tabular}{|c|c|c|>{\columncolor[gray]{0.85}}c|c|c|c|}
 \hline
 0 &  0 & a0 & a1 & a2 & a3 & a4\\ 
 0 & a0 & a1 & a2 & a3 & 04 &  0\\
a0 & a1 & a2 & a3 & a4 &  0 &  0\\
 0 &  0 & b0 & b1 & b2 & b3 & b4\\
 0 & b0 & b1 & b2 & b3 & b4 &  0\\
b0 & b1 & b2 & b3 & b4 &  0 &  0\\
 0 &  0 & c0 & c1 & c2 & c3 & c4\\
 0 & c0 & c1 & c2 & c3 & c4 &  0\\
c0 & c1 & c2 & c3 & c4 &  0 & 0\\
 \hline
\end{tabular}
\caption{Sub-part of column patch matrix of $3 \times 3$ patch size. The shaded column corresponds to the shaded patch in Table \ref{tab:2d-input}.}
\label{tab:2d-patch-matrix}
\end{table}
}

\snip{
\begin{table}[]
  \centering
   \setlength\tabcolsep{2pt}
\begin{tabular}{|c|c|c|c|c|}
\hline
a0 & \cellcolor[gray]{0.85} a1 & \cellcolor[gray]{0.85} a2 & \cellcolor[gray]{0.85} a3 & a4 \\
b0 & \cellcolor[gray]{0.85}b1 & \cellcolor[gray]{0.85}b2 & \cellcolor[gray]{0.8}b3 & b4 \\
c0 & \cellcolor[gray]{0.85}c1 & \cellcolor[gray]{0.85}c2 & \cellcolor[gray]{0.85}c3 & c4 \\
d0 & d1 & d2 & d3 & d4 \\
\hline
\end{tabular}
\caption{2D sub-part of input matrix. Shaded area is a $3 \times 3$ patch that is copied to a single column of the patch matrix in Table \ref{tab:2d-patch-matrix}.}
\label{tab:2d-input}
\end{table}
}

\begin{figure*}[t!]
	\centering
	\includegraphics[width=\linewidth]{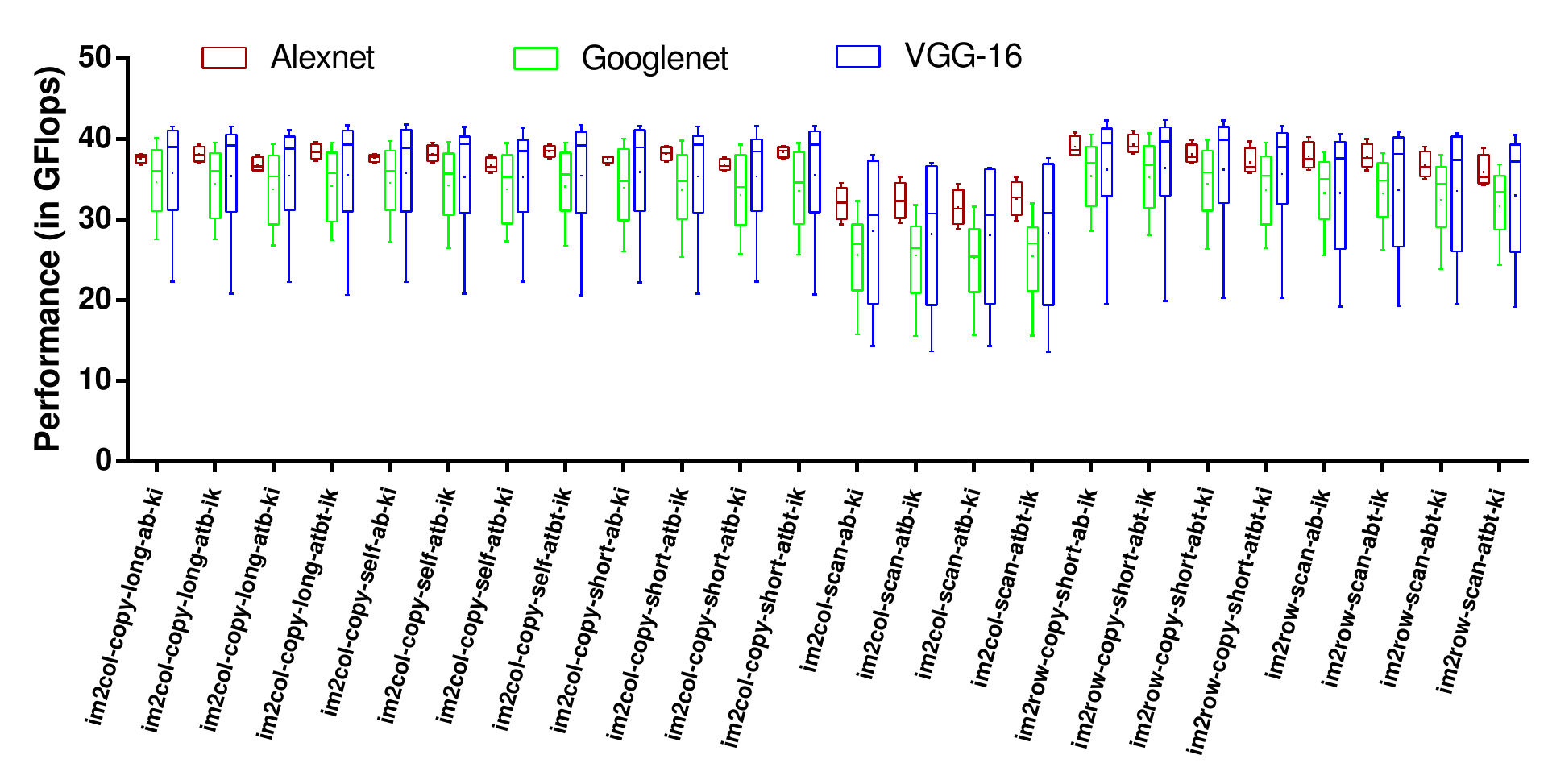}
	\caption{Performance of \gls{gemm}-based convolution with $O(K^2CHW)$ patch matrix}
	\label{fig:ksqu-im2-perf}
\end{figure*}

\subsection{Patch-building Algoritms}
\label{sec:patc-buil-algo}

\gls{i2c-scan}: The \gls{i2c-scan} method uses a patch building procedure that is very similar to the method outlined in \ref{fig:im2row-code} where each $KKC$ column of the patch matrix is built sequentially. While this method is simple to understand and can produce strided and unstrided patch matrix it has very poor spatial locality while accessing the input as each column of the patch matrix will use values from a number of the input's rows and columns.  

\gls{i2c-self}: The \gls{i2c-self} method is an improvement on the \gls{i2c-scan} method. From \ref{tab:2d-patch-matrix} we can see that many of the rows share large contiguous sections of the same values. With this we only need to use \gls{i2c-scan} to build a minority of the rows and then can use memory copying functions to construct the other rows of the patch matrix much faster.

\gls{i2c-long}: Looking at \ref{tab:2d-patch-matrix} we can see that the rows of the patch matrix are built from contiguous sections of the input (given a row-major input). With this we can build the patch matrix more quickly by copying entire sections of the input into the patch matrix. Note however that this property does not hold with \ref{tab:strided-patch-matrix} so this method can not be used for strided patch matrix.

\gls{i2c-short}: \gls{i2c-short} is very similar to \gls{i2c-long}. The difference comes in the fact that while not shown in \ref{tab:2d-patch-matrix} some of the values in the large contiguous sections copied from the input to the patch matrix will be replaced by zeroes. In \gls{i2c-long} we copy the longest possible section and then go back replacing values with zeroes where needed. In \gls{i2c-short} we copy the values up to each zero, insert the zero, then copy the next section. Also note \gls{i2r-long} does not exist but, \gls{i2r-short} can create strided patch matrix unlike \gls{i2c-short}.

\subsection{Evaluation}
Figure~\ref{fig:ksqu-im2-perf} exhibits the trade-offs between the different data layouts and the patch building algorithms discussed in section~\ref{sec:patc-buil-algo}. The suffixes \texttt{-ab-ki} denote the type of \gls{gemm} used. For instance, the first part of suffix denotes the layout of the two matrices. \texttt{ab} is the default \gls{gemm}, while \texttt{atb}, \texttt{abt} and \texttt{atbt} correspond to the respective \gls{gemm} call. The second part of the suffix denotes the order of the multiplication. \texttt{ik} performs input times kernel while \texttt{ki} performs kernel times input. These suffix notations are used throughout the rest of the paper and hold the same meaning.

The results for \gls{alexnet}, \gls{googlenet} and \gls{vgg} are shown on the same graph as box and whiskers. This type of graph gives a lot of information about what are the operating characteristics of the methods in each network. The minimum and maximum performance delivered by the methods are shown by the whiskers, while the box shows the 5 and 95 percentiles while the dash in the middle of the box shows the 50$^{th}$ percentile. The average performance of the method is indicated by the small \texttt{+} inside the box.

It is fairly evident from the scales of the box and whiskers from the graph that most of the \textit{im2} methods deliver performance in a small band. This implies that the performance delivered by a method is similar for all the layers in alexnet thereby making the variance in performance low. On the other hand, the variance increases in \gls{googlenet} and is the highest in \gls{vgg}. Another observation is that all the \gls{i2c-scan} variants produce significantly lower performance compared to the rest of the \textit{im2} methods. This is in keeping with our intuition as the \gls{i2c-scan} patch building algorithm has the worst locality amongst all the patch building methods.

%% file: methodology.tex

\section{Convolution with $O(K^2MHW)$ result matrix}
\label{sec:im2-meth}

A major problem with the convolution methods from Section \ref{sec:im2c} is
that the patch matrix requires $K^2$ more memory than the original
image, which is a dramatic increase in the size of the input.  This
additional space may reduce data locality, increase memory traffic,
and may exceed the available memory in embedded systems.

Figure~\ref{fig:covCode} shows a simplified loop nest for $\kh \times \kw$ convolution with $\kn$ kernels each with $\ic$ channels. A common operation in \glspl{cnn} such a \gls{googlenet}~\cite{Szegedy:2015} is convolution with a set of $1 \times 1$ convolutions. If we consider the code in figure \ref{fig:covCode} for the case where $\kh=1$, then the $x$ and $y$ loops collapse into a single iteration.

The resulting code is equivalent to 2D matrix multiplication of a $\kn \times \kc$ kernel times a $[\ic] \times [\ih \times \iw]$ input which results in a $[\kn] \times [\ih \times \iw]$ output. This output however is actually $\kn$ planes of $\ih\times\iw$ pixels which corresponds to an output of size $[\ih]\times[\iw]$ and $\kn$ channels. Let us call this correspondence of a $[\kn] \times [\ih \times \iw]$ matrix to an output matrix of size $[\ih]\times[\iw]$ and $\kn$ channels its \emph{multi-channel representation}, which we will use throughout the rest of this section. In other words, $1 \times 1$ \gls{mcmk} can be implemented by simply calling \gls{gemm} without data replication.

\subsection{\gls{k2r} and \gls{k2c}}
\label{sec:k2r}


Given that we can compute $1 \times 1$ \gls{mcmk} without data replication, how can we implement $\kh \times \kw$ \gls{mcmk}, for $\kh > 1$? We argue that a $\kh \times \kw$ convolution can be expressed as the sum of $\kh^2$ separate $1 \times 1$ convolutions. However the sum is not trivial to compute. Each $1 \times 1$ convolution yields a result matrix with dimensions $[\kn] \times [\ih \times \iw]$. We cannot simply add each of the resulting matrices pointwise, as each resultant matrix corresponds to a different kernel value in the $\kh \times \kw$ kernel. The addition of these matrices can then be resolved by offsetting every pixel in every channel of the \emph{multi-channel representation} of these matrices, vertically and/or horizontally (row and column offsets) by one or more positions before the addition.

\begin{figure}[t]
	\centering
	\includegraphics[width=\linewidth]{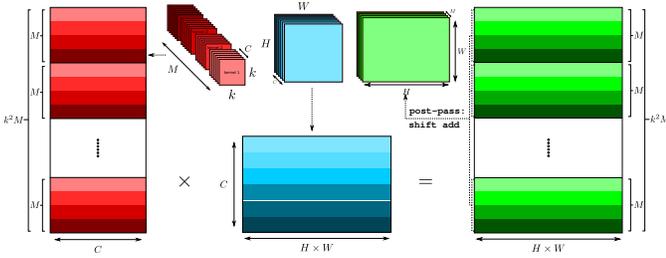}
	\caption{\gls{mcmk} using the ``\gls{k2r}'' method}
	\label{fig:k2r}
\end{figure}

For example, when computing a $3 \times 3$ convolution the result from computing the $1 \times 1$ \gls{mcmk} for the central point of the $3 \times 3$ kernel is perfectly aligned with the final sum matrix. On the other hand, the matrix that results from computing the $1 \times 1$ \gls{mcmk} for the upper left value of the $3 \times 3$ kernel must be offset up by one place and left by one place (in its \emph{multi-channel representation}) before being added to the final sum that computes the $3 \times 3$ \gls{mcmk}. Note that when intermediate results of $1 \times 1$ convolutions are offset, some values of the offsetted matrix fall outside the boundaries of the final result matrix. These out-of-bounds values are simply discarded when computing the sum of $1 \times 1$ convolutions.

It is possible to compute each of the $\kh^2$ separate $1 \times 1$ convolutions using a single matrix multiplication. We re-order the kernel matrix, so that the channel data is laid out contiguously, i.e. $\kn$ is the outer dimension and $\ic$ the inner. This data re-arrangement can be made statically ahead of time and used for all \gls{mcmk} invocations thereafter. Using a single call to \gls{gemm}, we multiply a $[\kh^2 \times \kn] \times [\ic]$ kernel matrix by a $[\ic] \times [\ih \times \iw]$ input matrix, resulting in a $[\kh^2 \times \kn] \times [\ih \times \iw]$ matrix. We perform a post pass of \texttt{\textbf{shift-add}} by summing each of the $\kn^2$ submatrices of size $\kn\times[\ih\times\iw]$ using appropriate offsetting in the \emph{multi-channel representation}. The result of this sum is a $[\kn] \times [\ih \times \iw]$ matrix, which is the output of our \gls{mcmk} algorithm. We refer to this as the \gls{k2r} algorithm.

If we swap the dimensions of the kernel matrix so that $\ic$ is not the innermost dimension and swap the input layout to make $\ic$ the innermost dimension, we get the \gls{k2c} algorithm. The \gls{gemm} call in this method would be to multiply an $[\ih \times \iw]\times[\ic]$ input matrix by a $[\ic]\times[\kh^2 \times \kn]$ kernel matrix, resulting in a $[\ih \times \iw]\times[\kh^2 \times \kn]$ matrix.

\begin{figure}[h]
	\centering
	\includegraphics[width=\linewidth]{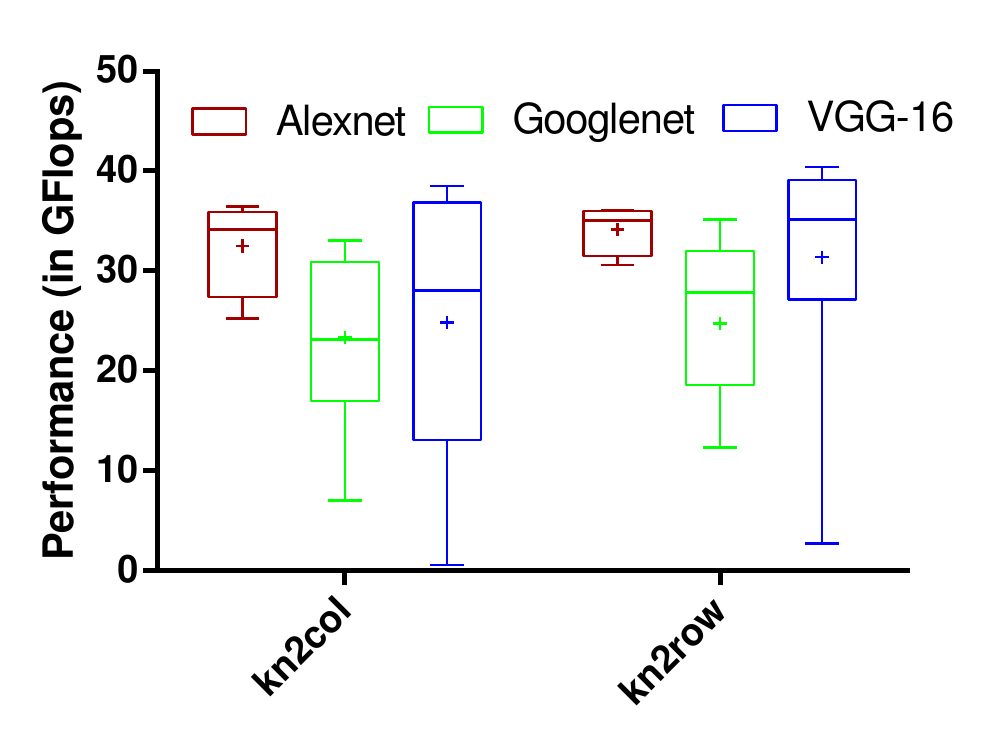}
	\caption{Performance of \gls{gemm}-based convolution with $O(K^2)$ data growth in result matrix}
	\label{fig:ksqu-kn2-perf}
\end{figure}

Similar to figure~\ref{fig:ksqu-im2-perf}, figure~\ref{fig:ksqu-kn2-perf} compares the performance of the two \textit{kn2} methods from Vasudevan et al.~\cite{Vasudevan:2017}. The \gls{k2r} method enjoys better locality and this is reflected in the performance delivered. The \gls{k2r} also produces a slightly tighter spread of performance across layers in all the networks, most notably in \gls{alexnet}. Another interesting observation is that the performance of both \gls{k2c} and \gls{k2r} tend to increase with the $\ic$ and $\kn$ and decreases with the size of the kernel.

\begin{figure*}[t]
\centering
\subfloat[Row-major linear layout of $11 \times 4$ matrix]{
  \label{fig:rm}
  \centering
  \setlength\tabcolsep{1pt}
  \begin{tabular}{|c|c|c|c|
                   c|c|c|c|
                   c|c|c|c|
                   c|c|c|c|
                   c|c|c|c|
                   c|c|c|c|
                   c|c|c|c|
                   c|c|c|c|
                   c|c|c|c|
                   c|c|c|c|
                   c|c|c|c|}
 \hline
\cellcolor[gray]{0.7}a0 & \cellcolor[gray]{0.7}a1 & \cellcolor[gray]{0.7}a2 & \cellcolor[gray]{0.7}a3 &
\cellcolor[gray]{0.91}b0 & \cellcolor[gray]{0.91}b1 & \cellcolor[gray]{0.91}b2 & \cellcolor[gray]{0.91}b3 &
\cellcolor[gray]{0.6}c0 & \cellcolor[gray]{0.6}c1 & \cellcolor[gray]{0.6}c2 & \cellcolor[gray]{0.6}c3 &
\cellcolor[gray]{0.8}d0 & \cellcolor[gray]{0.8}d1 & \cellcolor[gray]{0.8}d2 & \cellcolor[gray]{0.8}d3 &
 \cellcolor[gray]{0.5}e0 & \cellcolor[gray]{0.5}e1 & \cellcolor[gray]{0.5}e2 & \cellcolor[gray]{0.5}e3 &
 \cellcolor[gray]{0.7}f0 & \cellcolor[gray]{0.7}f1 & \cellcolor[gray]{0.7}f2 & \cellcolor[gray]{0.7}f3 &
 \cellcolor[gray]{0.91}g0 & \cellcolor[gray]{0.91}g1 & \cellcolor[gray]{0.91}g2 & \cellcolor[gray]{0.91}g3 &
 \cellcolor[gray]{0.6}h0 & \cellcolor[gray]{0.6}h1 & \cellcolor[gray]{0.6}h2 & \cellcolor[gray]{0.6}h3 &
 \cellcolor[gray]{0.8}i0 & \cellcolor[gray]{0.8}i1 & \cellcolor[gray]{0.8}i2 & \cellcolor[gray]{0.8}i3 &
 \cellcolor[gray]{0.95}j0 & \cellcolor[gray]{0.95}j1 & \cellcolor[gray]{0.95}j2 & \cellcolor[gray]{0.95}j3 &
 \cellcolor[gray]{0.7}k0 & \cellcolor[gray]{0.7}k1 & \cellcolor[gray]{0.7}k2 & \cellcolor[gray]{0.7}k3 \\
 \hline
\end{tabular}
}

\subfloat[The same data in memory reimagined in three different ways as $3 \times 12$ matrices]{
  \label{fig:sub-reimag}
  \centering
  \setlength\tabcolsep{1pt}
  \begin{tabular}{|c|c|c|c|
                   c|c|c|c|
                   c|c|c|c|
                   c|c|c|c|
                   c|c|c|c|
                   c|c|c|c|
                   c|c|c|c|
                   c|c|c|c|
                   c|c|c|c|
                   c|c|c|c|
                   c|c|c|c|}
 \hline
\cellcolor[gray]{0.91}a0 & \cellcolor[gray]{0.91}a1 & \cellcolor[gray]{0.91}a2 & \cellcolor[gray]{0.91}a3 &
\cellcolor[gray]{0.91}b0 & \cellcolor[gray]{0.91}b1 & \cellcolor[gray]{0.91}b2 & \cellcolor[gray]{0.91}b3 &
\cellcolor[gray]{0.91}c0 & \cellcolor[gray]{0.91}c1 & \cellcolor[gray]{0.91}c2 & \cellcolor[gray]{0.91}c3 &
\cellcolor[gray]{0.5}d0 & \cellcolor[gray]{0.5}d1 & \cellcolor[gray]{0.5}d2 & \cellcolor[gray]{0.5}d3 &
 \cellcolor[gray]{0.5}e0 & \cellcolor[gray]{0.5}e1 & \cellcolor[gray]{0.5}e2 & \cellcolor[gray]{0.5}e3 &
 \cellcolor[gray]{0.5}f0 & \cellcolor[gray]{0.5}f1 & \cellcolor[gray]{0.5}f2 & \cellcolor[gray]{0.5}f3 &
 \cellcolor[gray]{0.8}g0 & \cellcolor[gray]{0.8}g1 & \cellcolor[gray]{0.8}g2 & \cellcolor[gray]{0.8}g3 &
 \cellcolor[gray]{0.8}h0 & \cellcolor[gray]{0.8}h1 & \cellcolor[gray]{0.8}h2 & \cellcolor[gray]{0.8}h3 &
 \cellcolor[gray]{0.8}i0 & \cellcolor[gray]{0.8}i1 & \cellcolor[gray]{0.8}i2 & \cellcolor[gray]{0.8}i3 &
 j0 & j1 & j2 & j3 &
 k0 & k1 & k2 & k3 \\
 \hline
a0 & a1 & a2 & a3 &
\cellcolor[gray]{0.91}b0 & \cellcolor[gray]{0.91}b1 & \cellcolor[gray]{0.91}b2 & \cellcolor[gray]{0.91}b3 &
\cellcolor[gray]{0.91}c0 & \cellcolor[gray]{0.91}c1 & \cellcolor[gray]{0.91}c2 & \cellcolor[gray]{0.91}c3 &
\cellcolor[gray]{0.91}d0 & \cellcolor[gray]{0.91}d1 & \cellcolor[gray]{0.91}d2 & \cellcolor[gray]{0.91}d3 &
 \cellcolor[gray]{0.5}e0 & \cellcolor[gray]{0.5}e1 & \cellcolor[gray]{0.5}e2 & \cellcolor[gray]{0.5}e3 &
 \cellcolor[gray]{0.5}f0 & \cellcolor[gray]{0.5}f1 & \cellcolor[gray]{0.5}f2 & \cellcolor[gray]{0.5}f3 &
 \cellcolor[gray]{0.5}g0 & \cellcolor[gray]{0.5}g1 & \cellcolor[gray]{0.5}g2 & \cellcolor[gray]{0.5}g3 &
 \cellcolor[gray]{0.8}h0 & \cellcolor[gray]{0.8}h1 & \cellcolor[gray]{0.8}h2 & \cellcolor[gray]{0.8}h3 &
 \cellcolor[gray]{0.8}i0 & \cellcolor[gray]{0.8}i1 & \cellcolor[gray]{0.8}i2 & \cellcolor[gray]{0.8}i3 &
  \cellcolor[gray]{0.8}j0 &  \cellcolor[gray]{0.8}j1 &  \cellcolor[gray]{0.8}j2 &  \cellcolor[gray]{0.8}j3 &
 k0 & k1 & k2 & k3 \\
 \hline
a0 & a1 & a2 & a3 &
b0 & b1 & b2 & b3 &
\cellcolor[gray]{0.91}c0 & \cellcolor[gray]{0.91}c1 & \cellcolor[gray]{0.91}c2 & \cellcolor[gray]{0.91}c3 &
\cellcolor[gray]{0.91}d0 & \cellcolor[gray]{0.91}d1 & \cellcolor[gray]{0.91}d2 & \cellcolor[gray]{0.91}d3 &
 \cellcolor[gray]{0.91}e0 & \cellcolor[gray]{0.91}e1 & \cellcolor[gray]{0.91}e2 & \cellcolor[gray]{0.91}e3 &
 \cellcolor[gray]{0.5}f0 & \cellcolor[gray]{0.5}f1 & \cellcolor[gray]{0.5}f2 & \cellcolor[gray]{0.5}f3 &
 \cellcolor[gray]{0.5}g0 & \cellcolor[gray]{0.5}g1 & \cellcolor[gray]{0.5}g2 & \cellcolor[gray]{0.5}g3 &
 \cellcolor[gray]{0.5}h0 & \cellcolor[gray]{0.5}h1 & \cellcolor[gray]{0.5}h2 & \cellcolor[gray]{0.5}h3 &
 \cellcolor[gray]{0.8}i0 & \cellcolor[gray]{0.8}i1 & \cellcolor[gray]{0.8}i2 & \cellcolor[gray]{0.8}i3 &
 \cellcolor[gray]{0.8}j0 & \cellcolor[gray]{0.8}j1 & \cellcolor[gray]{0.8}j2 & \cellcolor[gray]{0.8}j3 &
 \cellcolor[gray]{0.8}k0 & \cellcolor[gray]{0.8}k1 &\cellcolor[gray]{0.8}k2 & \cellcolor[gray]{0.8}k3 \\
 \hline
\end{tabular}
}
\caption{A row-major $11 \times 4$ matrix and three different ways
  to reimagine sub-ranges of the matrix as $3 \times 12$ matrices.
  Note that the jagged parts at each end correspond to partial
  $3 \times 12$ matrices. To perform convolution with these partial
  matrices, zero-padding must be added to each end of the original
  matrix.
}
\label{fig:reimagining}
\end{figure*}

\begin{figure}[h]
	\centering
	\includegraphics[width=\linewidth]{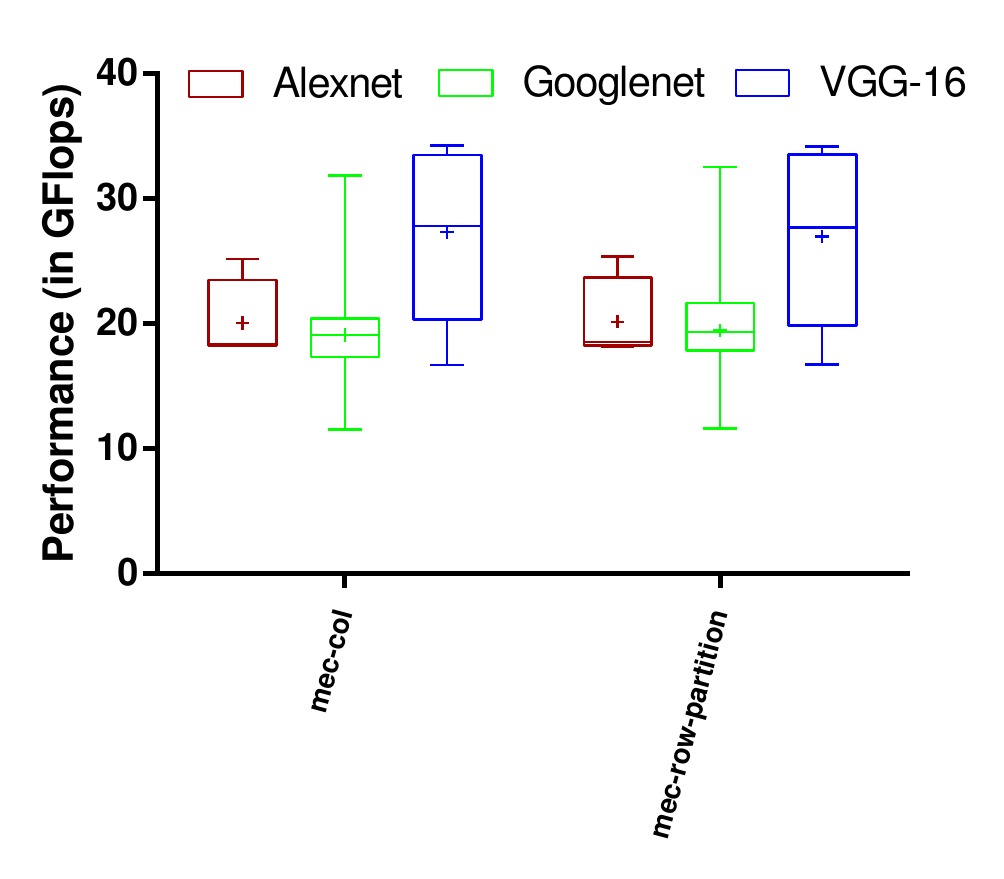}
	\caption{Performance of \gls{gemm}-based convolution with $O(K)$ data growth}
	\label{fig:k-all-perf}
\end{figure}

\section{GEMM-based Convolution with $O(KCHW)$ patch matrix}

GEMM-based convolution can make very good use of the underlying
hardware, but the method described in Section \ref{sec:im2c} requires a patch
matrix with a factor of $K^2$ increase in the size of the input, and
Section \ref{sec:im2-meth} requires a $K^2$ increase in the output of the GEMM.
In this section we describe an existing approach that requires just a factor
of $K$ increase in the size of the input of the GEMM.
This reduction in memory size is particularly welcome on
memory-limited embeddes systems, but it can also improve locality and
reduce memory traffic on other systems. Reducing memory traffic can be
particularly important for multicore systems where all cores share a
single interface to off-chip main memory.

\subsection{Reimagining matrix dimensions}
Memory-Efficient Convolution (MEC)~\cite{mec-paper} is an extension of
the classical \gls{i2c} which does not require a full $O(K^2)$ patch
matrix. The core idea of the algorithm relies on the layout of
matrices in memory.

In C/C++ elements of a 1D array are contiguous in memory. A 2D array
consists of a sequence of 1D arrays that are laid out contiguously in
memory. Therefore, a 2D array with dimensions $H \times W$ has exactly
the same layout in memory as a 1D array of size $HW$. Thus, we can
switch between the two different interpretations of the data in memory
by simply changing the type of the pointer to the array, without
having to change the data in memory. It is this basic insight that
allows the 2D GEMM algorithm to operate on multidimensional tensors in
the all the algorithms presented in this paper.

Figure \ref{fig:reimagining} shows a 2D $11 \times 4$ matrix laid out
contiguously in memory. It is possible to reinterpret a sub-region of
this matrix as, say, a $3 \times 12$ matrix without changing the
layout in memory. By selecting different sub-regions, one can select
different $3 \times 12$ sub-matrices of the original. Using this
mechanism, we can perform 1D convolution by selecting a sequence of
$K$ rows of the matrix and treating them as a single row.  The MEC
algorithm \cite{mec-paper} uses this approach to avoid having to
create a patch matrix that is $O(K^2)$ times the size of the original
matrix. Instead MEC performs 1D convolutions by performing multiple
GEMM operations on overlapping regions of the matrix. The MEC approach
allows horizontal 1D convolutions to be performed without having to
create a patch matrix, but the same trick cannot be used for vertical
1D convolutions, or indeed 2D convolutions.

To overcome this problem, the MEC algorithm creates a patch matrix
that is $O(K)$ times the size of the original input to allow the
vertical part of convolutions, and uses performs overlapping GEMMs for
the horizontal part. The result is an algorithm that requires no
more arithmetic operations than the classical \gls{i2c} algorithm,
but requires only a factor of $O(K)$ times the input size in
additional memory, rather than the $O(K^2)$ required by \gls{i2c}.

\section{GEMM-based convolution with sub-$K$ data growth}
\label{sec:aramckmk}
The \gls{k2r} algorithm eliminates the need for data replication in the input (\gls{i2c} requires a $\kh^2$ expansion of the input). However this results in a $\kh^2$ expansion in the result of the matrix multiplication. This problem might be thought of like a tube of toothpaste; if we squeeze the $\kh^2$ data expansion from the input image, it re-appears in the output of the matrix multiplication. 


\subsection{Accumulating \gls{k2r} and \gls{k2c}}
\label{sec:k2r-as}

\begin{figure}[h]
	\centering
	\includegraphics[width=\linewidth]{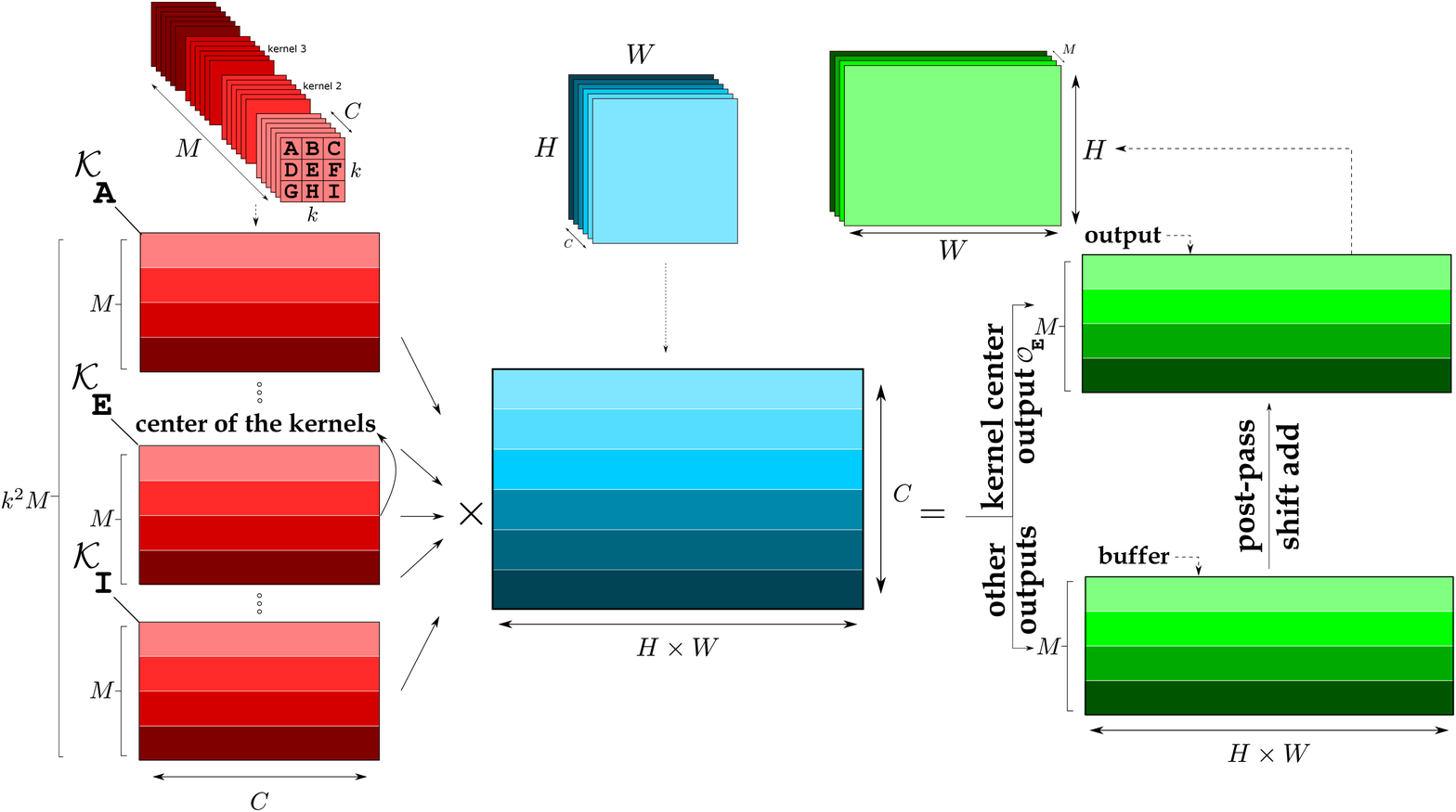}
	\caption{\gls{mcmk} using the ``\gls{k2r-as}'' method}
	\label{fig:k2r-as}
\end{figure}

In order to avoid this ``tube of toothpaste'' problem, we introduce two new methods which are accumulating variants of \gls{k2r} and \gls{k2c} called \gls{k2r-as} and \gls{k2c-as} respectively. Instead of treating the kernel as one big kernel matrix of size $[\kh^2 \times \kn] \times [\ic]$, the k2r-as accumulating method treats it as $\kh^2$ smaller $[\kn\times\ic]$ sized kernel matrices ($\mathcal{K}_{A}\cdots\mathcal{K}_{I}$) as shown in the Figure~\ref{fig:k2r-as}.

\snip{
\begin{figure*}[t!]
	\centering
  \subfloat[VGG-16]{
    \centering
    \includegraphics[scale=0.5]{figures/graphs/A57/vgg-16}
    \label{fig:A57-vgg}
  }\qquad
  \subfloat[Googlenet]{
    \centering
    \includegraphics[scale=0.5]{figures/graphs/A57/googlenet}
    \label{fig:A57-googlenet}
  }
  \subfloat[Alexnet]{
    \centering
    \includegraphics[scale=0.5]{figures/graphs/A57/alexnet}
    \label{fig:A57-alexnet}
  }
	\caption{Execution time for selected layers of \gls{googlenet}, \gls{vgg} and \gls{alexnet} on the ARM Cortex A57.\textbf{Lower is better}. }
	\label{fig:A57-summary}
\end{figure*}

\begin{figure*}[h]
	\centering
  \subfloat[VGG-16]{
    \centering
    \includegraphics[scale=0.5]{figures/graphs/x86_64/vgg-16}
    \label{fig:x86_64-vgg}
  }\qquad
  \subfloat[Googlenet]{
    \centering
    \includegraphics[scale=0.5]{figures/graphs/x86_64/googlenet}
    \label{fig:x86_64-googlenet}
  }
  \subfloat[Alexnet]{
    \centering
    \includegraphics[scale=0.5]{figures/graphs/x86_64/alexnet}
    \label{fig:x86_64-alexnet}
  }
	\caption{Execution time for selected layers of \gls{googlenet}, \gls{vgg} and \gls{alexnet} on the Intel Core i5-3450. \textbf{Lower is better}. }
	\label{fig:x86_64-summary}
\end{figure*}
}

We then multiply the kernel $\mathcal{K}_{E}$ which corresponds to the center value of every channel of every kernel with the image matrix (shown in blue in the figure) of size $[\ic]\times[\ih\times\iw]$ which results in an output matrix ($\mathcal{O}_{E}$) of size $[\kn]\times[\ih\times\iw]$ which is stored in \texttt{\textbf{output}} directly without any shifting (as discussed in the section~\ref{sec:k2r}).

Kernel matrix $\mathcal{K}_{A}$ is then multiplied with the input matrix and the result is stored in \texttt{\textbf{buffer}}. Since the kernels in the figure are $3\times3$, the row offset for the outputs produced by $\mathcal{K}_{A}$ is $+1$ and the corresponding column offset is also $+1$ as $E$ is one row down and column to the right of $A$. The values in \texttt{buffer} are shifted by the calculated offsets in its \emph{multi-channel representation}. This process is repeated for the rest of the kernels ($\{\mathcal{K}_{B}\cdots\mathcal{K}_{I}\}-\{\mathcal{K}_{E}\}$). For a kernel of size $\kh\times\kw$ appropriate offset values are by calculating how far each value ($\mathcal{K}_{A}\cdots\mathcal{K}_{k^2}$) is away from the center of the kernel. Our \gls{k2c-as} method works in the same way as \gls{k2r-as}, but with the input and kernel matrices transposed.

Figure~\ref{fig:k2r-as} and its explanation in this section hold true for \gls{k2c-as} along with applicable transformations to the kernels, inputs and outputs (as we have explained for converting \gls{k2r} to \gls{k2c}).

\subsection{Accumulating with \gls{gemm}}
\label{sec:accu-with-gemm}

\begin{figure}[h]
	\centering
	\includegraphics[width=\linewidth]{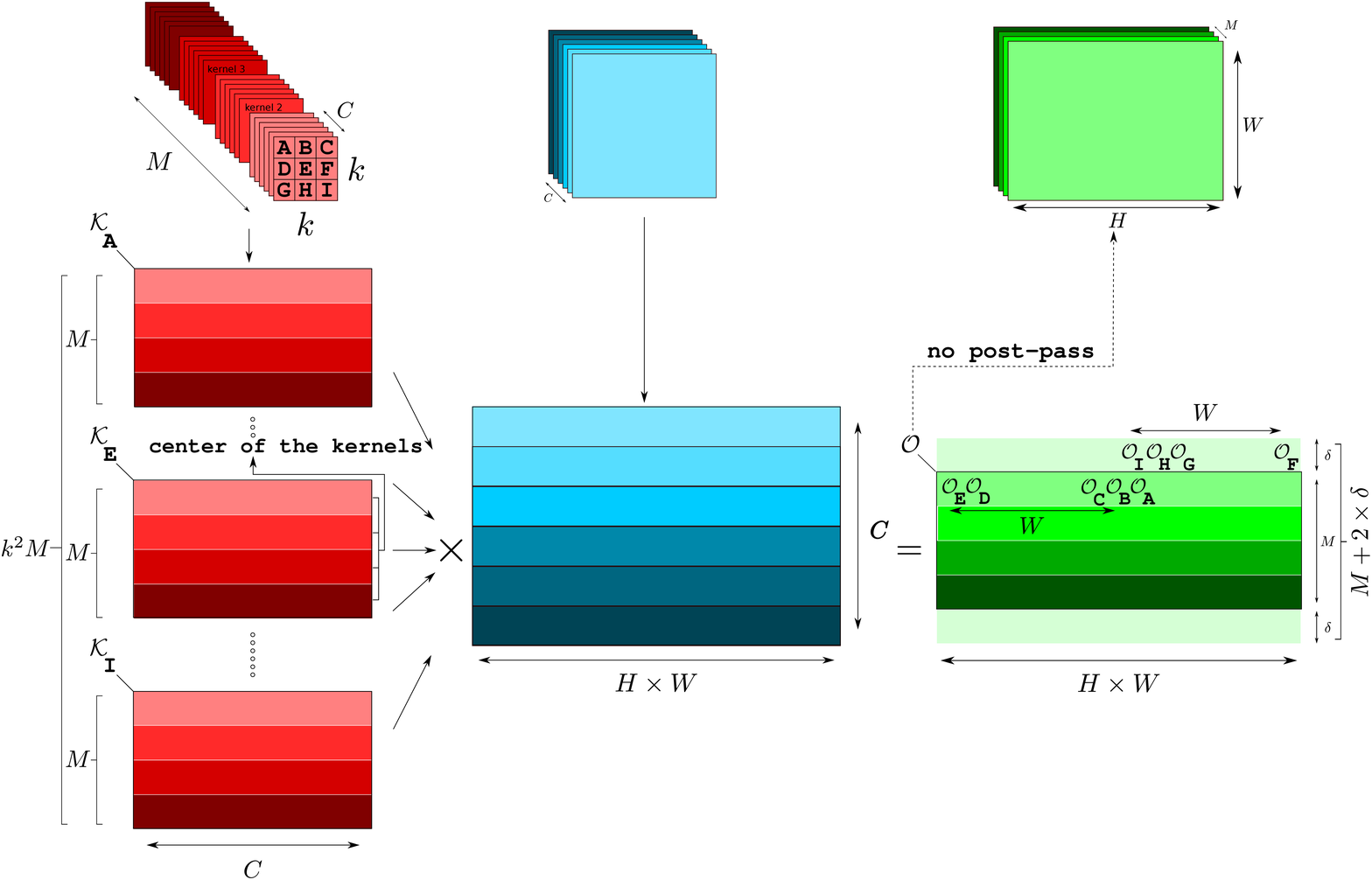}
	\caption{MCMK convolution using the accumulation that is built into the GEMM library call}
	\label{fig:k2r-aa}
\end{figure}
The signature of \gls{gemm} in most BLAS libraries is $C=\alpha\times(A\ast B)+\beta\times C$. One can transform a \gls{gemm} call into an output accumulating \gls{gemm}, i.e. multiply the two input matrices $A$ and $B$ and add it to the resultant matrix $C$, by setting the value of the scalar parameter $\beta$ to zero. The \textbf{k2r-aa} method, leverages this output accumulating feature of \gls{gemm} as shown in Figure~\ref{fig:k2r-aa}.

It works in the same way as \gls{k2r-as} by treating the big kernel matrix of size $[\kh^2 \times \kn] \times [\ic]$ as $\kh^2$ $[\kn\times\ic]$ smaller kernels ($\mathcal{K}_{A}\cdots\mathcal{K}_{I}$). To store the output, we reserve a contiguous piece of memory that is of size $[\kn+2\delta]\times[\ih\times\iw]$ where $\delta$ is the number of extra rows in the output matrix given by $\delta=\ceil{\frac{\kh}{2\ih}}$ for square kernel matrices.

Note that in the example in Figure \ref{fig:k2r-aa} padding is added to the top and bottom of the result matrix which can be written to by the GEMM calls. The values computed in these padding array elements are never actually used. In the example, we have assumed that we always allocate full rows of size $HW$ for padding. If, however, we only allocate just enough array elements that are needed for padding rather than allocating full rows, then the additional space needed for padding is $O(KW)$.



A \gls{gemm} of $\mathcal{K}_{E}$ (corresponding to the center value of every channel of every kernel) and the input matrix produces an output matrix $\mathcal{O}_{E}$ as in Figure~\ref{fig:k2r-aa}. Since this output corresponds to the center of the kernel, it does not have to be offset in the final output. If one were to add a similar output matrix $\mathcal{O}_{D}$ to the intermediate output then $\mathcal{O}_{D}$ has to be shifted to the right by one element as $\mathcal{D}$ is one element to the left of the center of the kernel. By the same logic, $\mathcal{O}_{F}$ has to be shifted left by one element as $\mathcal{F}$ is one element to the right of the center.

Thus, an element $\mathcal{X}$ in the kernel that is $x_1$ rows above the center and $y_1$ to the left of the center must be offset by $x_1\times\iw+y_1$. So we choose the starting location of the output associated with the first kernel ($\mathcal{K}_{A}$) such that the output corresponding to the center of the kernel will start at the start of the $\delta^th$ row of the intermediate output. This starting address is given by $\mathcal{O}_{A}=\mathcal{O}_{t}+\floor{\frac{\kh}{2}}(\ih\times\iw+\iw+1)$. Based on $\mathcal{O}_{A}$ and $\mathcal{O}_{E}$, we can calculate the rest of the starting addresses and supply them to successive accumulating \gls{gemm} calls. After $\kh^2$ \gls{gemm} calls have been made, the output starting at $\mathcal{O}_{E}$ denoted by $\mathcal{O}$ in Figure~\ref{fig:k2r-aa} contains the final output with correct accumulation of kernel and weight pairs.

Note that the final output $\mathcal{O}$ is a matrix of size $[\kn]\times[\ih\times\iw]$ starting at location $\mathcal{O}_{E}$. Therefore the edge pixels of the output are approximate values. At the boundaries of an image a convolution typically assumes that pixels outside of the image have a value of zero, but our \gls{k2r-aa} method instead takes the value from the opposite edge of the image.

\glsunset{k2r-aa}
\begin{figure*}[t!]
	\centering
	\includegraphics[width=\linewidth]{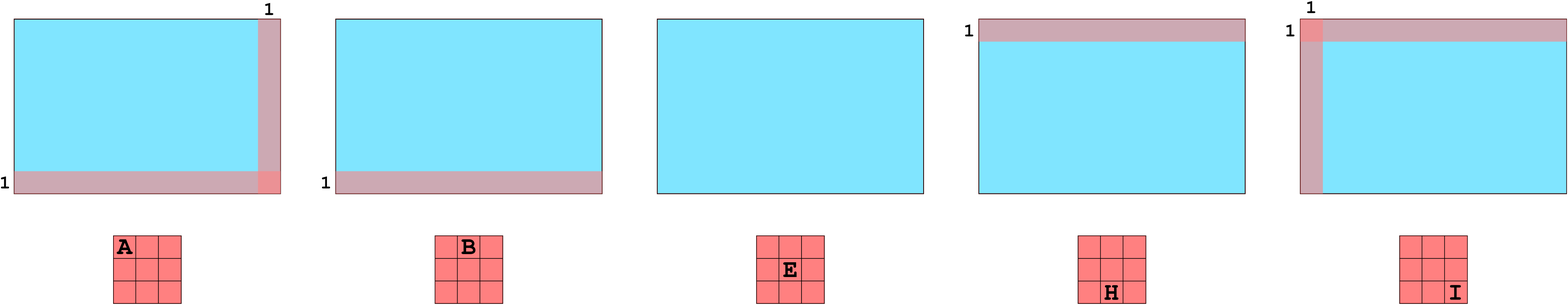}
	\caption{Hole punching in \gls{k2r-aa}}
	\label{fig:hole-punc}
\end{figure*}

One way to fix this is by applying a post-pass where a loop based implementation of \gls{mcmk} is applied to every erroneous error pixel. However, this has a significant impact on performance. In order to overcome the issue of producing correct results while being performant, we analyzed the pairs of products whose sum gave rise to the edge pixels in question.

Even though the right pairs of image pixels and kernel values are multiplied together, incorrect pairs are summed up together in the boundary pixels. As mentioned earlier, where image pixels are to be treated as zero (boundary handling) our \gls{k2r-aa} reads the pixel values from the subsequent row in the opposite edge of the image due to the contiguous memory layout.


In order to avoid summing incorrect pairs of image pixels and kernel values, we propose an intermediate step between every \gls{gemm} call, that multiplies the an $\kh \times \kc$ kernel matrix (that corresponds to all the kernel values for a given position in a $\kh \times \kw$ kernel) with the $\ic \times [\ih \times \iw]$ image matrix. In this intermediate step we ``punch holes" in the image matrix by setting certain positions of the image matrix to zero as shown in algorithm~\ref{fig:hole-punc}. The figure shows the regions that are zeroed out in a single channel of the input matrix for ease of understanding, but these regions are applicable for all channels in the input.

As discussed earlier, the intuition behind zeroing certain regions of the image before every accumulating GEMM call is to avoid unnecessary pairs of products being summed up together. For instance let's consider the first part of figure~\ref{fig:hole-punc} which corresponds to the GEMM of the input with the smaller kernel matrix ($\mathcal{K}_{A}$) corresponding to the \textbf{A}$^{th}$ element. If convolution is performed with this element placed on a pixel of the input that is in the marked region, the center of the kernel falls outside the boundaries of the output thereby making this product of input and kernel unnecessary for the end result.

Similarly, for the \textbf{B}$^{th}$ element, the last row of the input must be zeroed out. Note however, for the central element, no pixels need to be punched out because placing the \textbf{E}$^{th}$ element on any pixel of the input will always result in the output of that convolution inside the boundary of the output. Between every \gls{gemm} call, we restore the pixels that were saved before the previous \gls{gemm} and save the pixels where ``holes" are to be punched for the current \gls{gemm} call.

\begin{figure}[h]
	\centering
	\includegraphics[width=\linewidth]{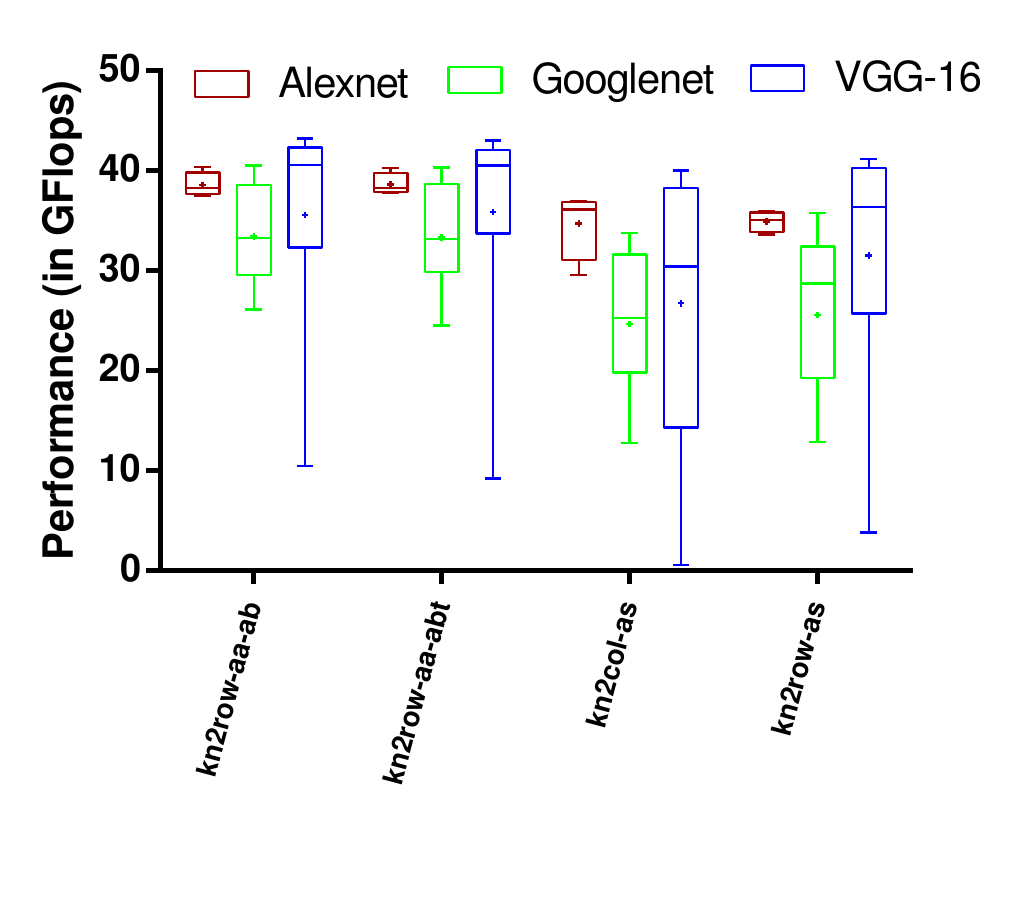}
	\caption{Performance of \gls{gemm}-based convolution with $O(<K)$ data growth}
	\label{fig:subk-all-perf}
\end{figure}

Figure~\ref{fig:subk-all-perf} compares the performance of the accumulating versions of \gls{k2r} and \gls{k2c} described in section~\ref{sec:k2r-as} and the accumulating \gls{gemm} version with hole punching described in this section. The accumulating \gls{k2r} (\gls{k2r-as}) and \gls{k2c} (\gls{k2c-as}) perform much better compared to their baseline non-accumulating versions described in section~\ref{sec:im2-meth}. However, the two variants of the accumulating \gls{gemm} based hole-punching method harness the output data locality offered by having \gls{gemm} accumulate into offsetted output locations and seem to outperform the other accumulating methods.

%% file: results.tex



\section{Results}
\label{sec:res}
\subsection{Testing Environment}
We evaluated the execution time of every previously mentioned convolution algorithm
across a number of input dimensions. The input dimensions chosen were selected from
three popular CNN architectures: AlexNet~\cite{Krizhevsky:12}, VGG-16~\cite{Simonyan14c}, and \gls{googlenet}~\cite{Szegedy:2015}.
We chose these input dimensions as we believe they represent realistic inputs
that the algorithms will be expected to handle when used for practical applications.
We ran our experiments on two general-purpose processors, an Intel i7 and an ARM\textregistered
Cortex\textregistered A57 (on a NVIDIA JTX1 board). We evaluated the speeds of all the algorithms
executed on processors using both a single thread running on a single core, and using all the CPU
cores allowing up to 8 threads on the i7 and 4 threads on the A57. The execution times
measured the time needed to allocate and construct any intermediate data structures (e.g.
the patch matrices for i2c and the extra output space for k2r), GEMM invocations and
writing results to the output buffer. We did not measure the time needed to convert the
input or output feature map to a specified format (i.e. we treated algorithms that produced
outputs in $CxHxW$ format and $HxWxC$ format as equally valid).

On the i7 processor our algorithms were compiled using gcc 7.1.1 while on the A57
gcc 5.4.0 was used as it ships with the NVIDIA JTX1 board. We statically linked with the
OpenBLAS library to implement GEMM. We used OpenBLAS version 0.2.2. We used OpenBLAS' internal
threading model to multithread the GEMM calls.

\subsection{Memory Usage}

\begin{table*}[th]
	\centering
	\resizebox{0.99\linewidth}{!}{%
		\begin{tabular}{||c|c|c||} 
			\hline
			\textbf{Method} & \textbf{Label} & \textbf{Extra memory}\\
			\hline
			\glsdesc{i2r} & im2row & $(\kh^2-1)\times((\ih\iw)\times\ic)$\\
			\hline
			\glsdesc{i2c} & im2col & $(\kh^2-1)\times((\ih\iw)\times\ic)$\\
			\hline
			\glsdesc{k2r} & kn2row & $(\kh^2-1)\times(\kn\times(\ih\iw))$\\
			\hline
			\glsdesc{k2c} & kn2col & $(\kh^2-1)\times(\kn\times(\ih\iw))$\\
			\hline
			\glsdesc{k2r-as} & kn2row-as & $(\ih\times\iw)\times\kn$\\
			\hline
			\glsdesc{k2c-as} & kn2col-as & $\kn\times(\ih\times\iw)$\\
			\hline
			\glsdesc{k2r-aa} & kn2row-aa & $\kh \times \iw$\\
			\hline
		\end{tabular}
	}
	\caption{Extra memory required}
	\label{table:extr-memo}
\end{table*}
While evaluating our algorithms on the JTX1 board we found that for a number of our chosen
input dimensions the board ran out of memory intermittetly while executing our $O(K^2)$
algorithms. This occurred during the execution of the upper layers of VGG16 where we have
a input feature map which is large in all dimensions. This demonstrates an area where the
$O(K)$ and $O(<K)$ algorithms are particular necessary. The execution times for the problematic input
dimensions have been omitted from the ARM result tables as the times for many of the $O(K^2)$
algorithms could not be produced.

\subsection{Resulting Trends}

The general theme throughout the results we have observed points towards the fact that there is no one method to outperform them all. From this gamut of implementations of the convolution layer, there is seemingly no one method that is the pick of the lot. Instead, given a layer's operating characteristics (i.e, number of channels $\kc$, kernels $\kn$, size of the kernel $\kh$ and input height $\ih$ and width $\iw$) most methods form clusters in their performance characteristics.

\begin{figure}[h]
	\centering
	\includegraphics[width=\linewidth]{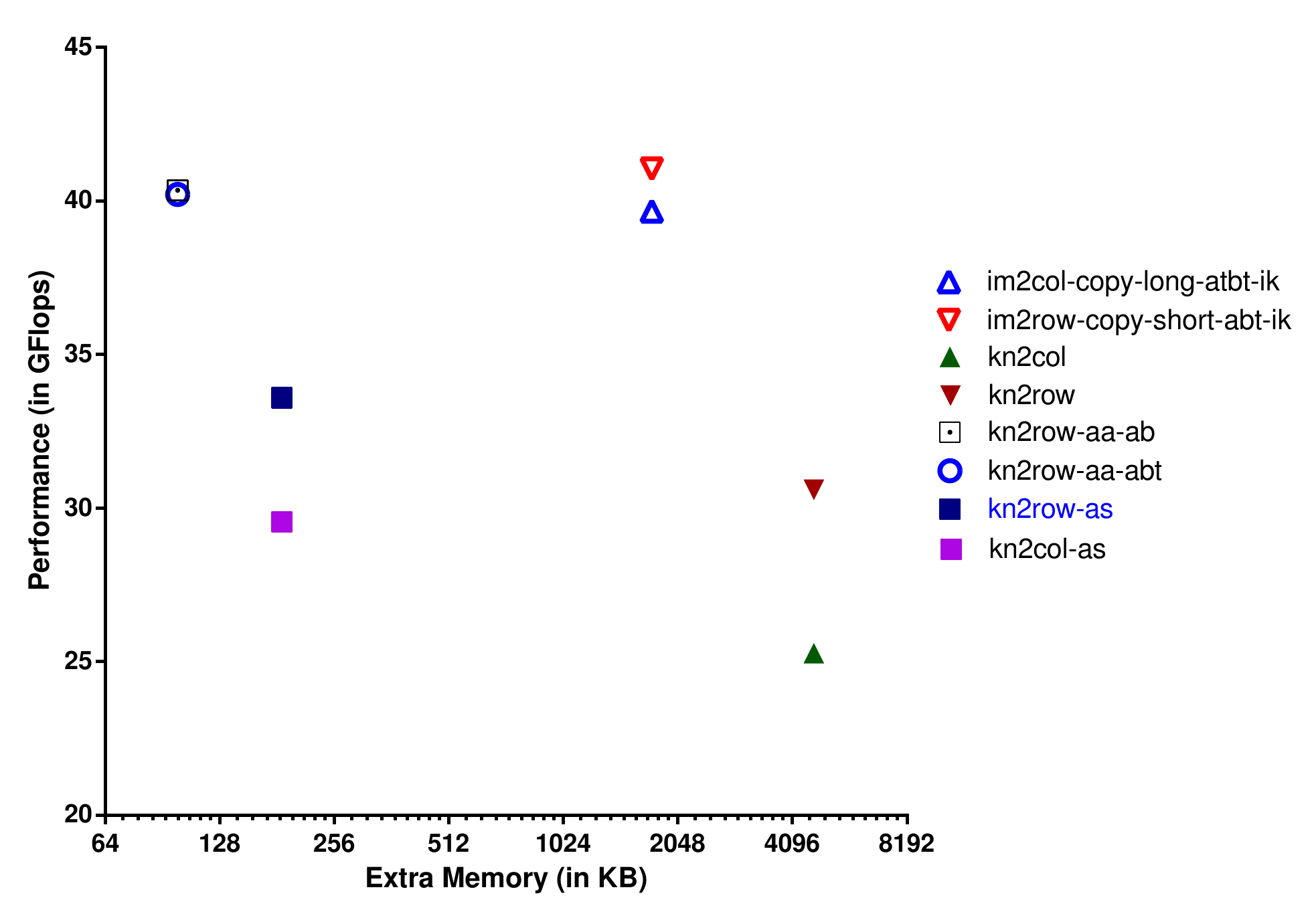}
	\caption{Performance vs extra memory required for all methods in conv2 of Alexnet}
	\label{fig:conv2-alex-perf-vs-data}
\end{figure}

\begin{figure}[h]
	\centering
	\includegraphics[width=\linewidth]{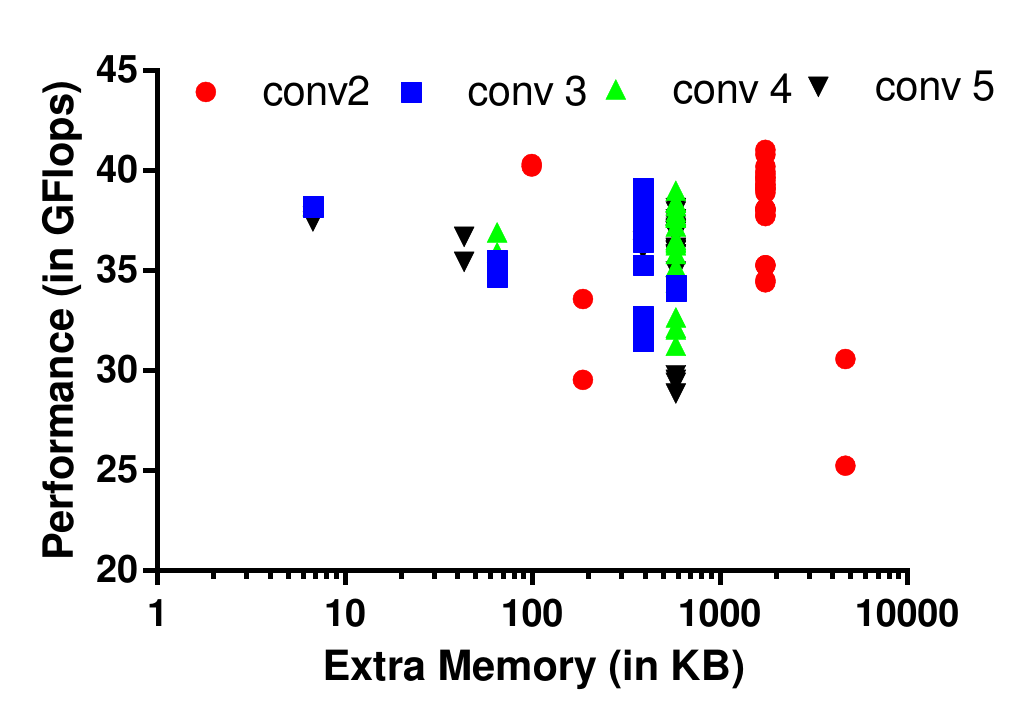}
	\caption{Performance vs extra memory required for all the methods in all layers of Alexnet}
	\label{fig:alex-perf-vs-data}
\end{figure}

This is further illustrated by figures~\ref{fig:conv2-alex-perf-vs-data} and~\ref{fig:alex-perf-vs-data}. Both these figures show the performance of a method compared against the extra memory required by each algorithm (in accordance with table~\ref{table:extr-memo}). Figure~\ref{fig:conv2-alex-perf-vs-data} is an instance of figure~\ref{fig:alex-perf-vs-data} as it shows the performance characteristics of selected methods for the second convolution layer of \gls{alexnet} while the latter represents all the methods and all the layers of \gls{alexnet}.

It is very evident from figure~\ref{fig:conv2-alex-perf-vs-data} that the non-accumulating \textit{kn2} methods require the most extra memory for the second convolution layer of \gls{alexnet} while providing low performance. However their accumulating counterparts require nearly $\dfrac{1}{20}^{th}$ the extra memory required while providing better runtimes. The \textit{im2} methods represented here are the best performing out of the gamut of methods presented in section~\ref{sec:im2-meth}. The hole-punching methods (\gls{k2r-aa}) are competent compared to the \textit{im2} methods while requiring the least amount of extra memory.


\begin{figure}[h]
	\centering
	\includegraphics[width=\linewidth]{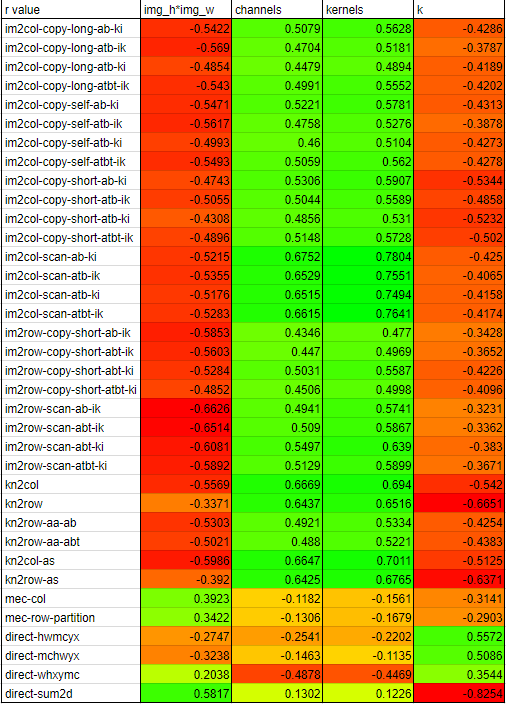}
	\caption{Correlation matrix of all methods vs convolution parameters}
	\label{fig:corr-matr-hori}
\end{figure}

Figure~\ref{fig:corr-matr-hori} shows the the correlation coefficient between each of the operating condition variables (i.e, number of channels $\kc$, kernels $\kn$, size of the kernel $k$ and input height times width $\ih \times \iw$). It is interesting to note that all the \gls{gemm} based methods have a positive correlation with the number of channels and number of kernels while they have a negative correlation with the number of pixels in the image and the size of the kernel. 

The non-accumulating \textit{kn2} methods have a worse negative correlation with the number of kernels, i.e. as the number of kernels tend to increase the performance tends to decrease which is explained by the fact that the size of the result matrix grows (along one direction; either rows or columns increase) as $k$ increases thereby resulting in a \gls{gemm} between oddly shaped matrices. The accumulating versions of the \textit{kn2} methods have an even worse effect as the number of distinct \gls{gemm} calls increases quadratically with $k$.

The $O(K)$ data growth methods have a positive correlation with the number of pixels in the input and a negative correlation with the number of channels and kernels. We have also added the direct loop based methods as described in figure~\ref{fig:covCode}. The suffixes for these methods indicate the loop ordering from the figure.

\begin{figure}[h]
	\centering
	\includegraphics[width=\linewidth]{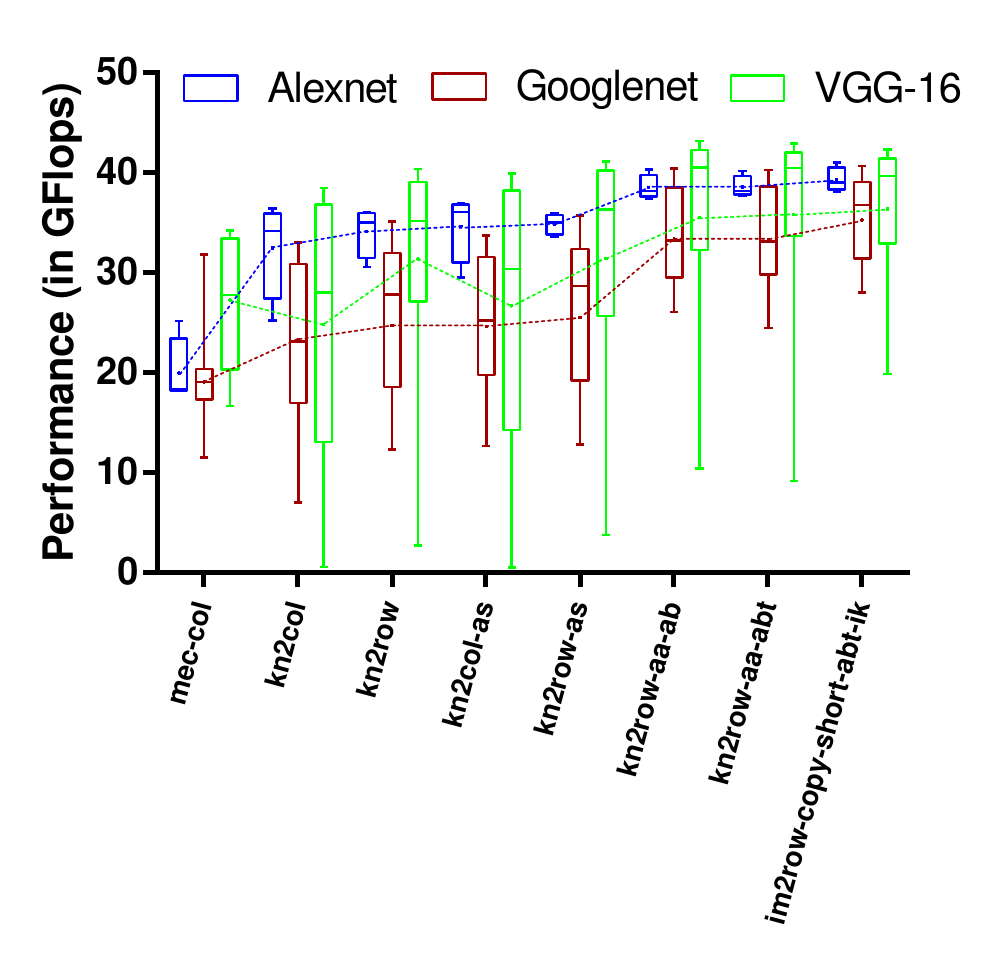}
	\caption{Performance of the best \gls{gemm}-based convolutions for \gls{alexnet}, \gls{googlenet} and \gls{vgg}}
	\label{fig:best-all-perf}
\end{figure}

These graphs clearly show that there is a wide range of performance characteristics displayed by the methods discussed in this paper. The ramifications of this are two-fold: (1) user's choice is paramount as the choice of implementations depend on the deployment environment, i.e. for a mobile based deployment scenario, the hole-punching methods are seemingly the correct choice as they require the least overhead in memory requirement and are competent with the other methods in terms of performance and (2) the optimal choice of implementations and layouts to minimize the total execution time of a network posits a complicated optimization problem.


\subsection{Other Experiments}
We also implemented versions of our algorithms that used OpenMP to parallelize the data
transformations needed to construct the intermediate data structures in \gls{i2c}, \gls{i2r} and
\gls{mec}. However we found that the speed improvements from this were negilible. We used OpenMP
4.0.1-1 and allowed eight threads on the i7 processor and 4 threads on the A57.

We also measured the number of stall cycles that occurred due to L1 and L3 cache while running our
tabled experiments on the i7 processor. We found that the optimal version of k2r (labeled kn2row-aa in
the tables) had on average the least number of stall cycles, although for the upper layers of VGG16
\gls{mec} fared best. This falls in line with the execution times recorded on the i7 processor.

%% file: related-work.tex


\section{Related Work}
\label{sec:rela-work}

The \gls{i2c} method of performing \gls{mcmk} is an extension of well-known
methods of performing 2D convolution using a Toeplitz matrix. Chellapilla et
al.~\cite{chellapilla2006high} are the first researchers to implement
\gls{mcmk} using using \gls{i2c}. They report significant speedups compared to
the simpler approach of summing multiple channels of 2D convolutions.


Yanqing et al. rediscovered \gls{i2c} for the Caffe deep learning
system~\cite{jia2014caffe}, which uses GPUs and other accelerators to speed up
\glspl{dnn}. The \gls{i2c} approach remains the most widely-used way to
implement \gls{mcmk}, and is used in deep learning frameworks such as Caffe,
Theano and Torch.


Gu et al. \cite{Gu:2016} apply \gls{i2c} to a batch input images to
create a column matrix for multiple input images. They find that
batching can improve throughput by better matching the input matrix
sizes to the optimal sizes for their \gls{gemm} library.


%

Tsai et al. \cite{Tsai:2016} present a set of configurable OpenCL kernels for
\gls{mcmk}. By coding the \gls{mcmk} loop nests directly they eliminate the
need for \gls{i2c} data replication, and thus allow the use of larger batch
sizes while maintaining constraints on local memory. The found that the
performance of a naive loop nest for \gls{mcmk} is not good, but they achieve
satisfactory performance with a program generator and autotuner.

Chetlur et al. \cite{Chetlur:2014} propose a \gls{gemm}-based approach to
convolution based on \gls{i2c}. However, rather than creating the entire column
matrix in one piece, they instead lazily create sub-tiles of the column matrix
in on-chip memory. To optimize performance, they match the size of their
sub-matrix tiles to the tile sizes used by the underlying \gls{gemm}
implementation. They find that this lazy \gls{i2c} achieves speedups over
Caffe's standard \gls{i2c} of between around 0\% and 30\%.

%% file: resultstable.tex
\setlength\extrarowheight{0pt}

\definecolor{cnnresultcolor1}{rgb}{1,0.0,0}
\definecolor{cnnresultcolor2}{rgb}{1,0.5,0}
\definecolor{cnnresultcolor3}{rgb}{1,1.0,0}
\clearpage
{ \footnotesize
{ \centering
{ \setlength{\tabcolsep}{0.007cm}
\begin{table*}[t]
\subfloat[single-threaded]{
		\begin{tabular}{ |c|c|c|c|c|c|c|c|c|c|c|c|c|c|c|c|c|c|c|c|c|c| }
			\hline
			height   &27&13&13&13&57&57&28&28&14&7&224&224&112&112&56&56&28&28&14&14   \\
			\hline
			width    &27&13&13&13&57&57&28&28&14&7&224&224&112&112&56&56&28&28&14&14   \\
			\hline
			channels &96&256&384&384&64&64&16&96&160&832&3&64&64&128&128&256&256&512&512& 1024   \\
			\hline
			k        &5&3&3&3&1&1&5&3&3&1&3&3&3&3&3&3&3&3&3&3   \\
			\hline
			kernels  &256&384&384&256&64&192&32&128&320&384&64&64&128&128&256&256&512&512&1024&1024     \\
			\hline
			\hline
			im2col-copy-long-ab-ki & 21.13 \cellcolor{white} & 7.135 \cellcolor{white} & 10.71 \cellcolor{white} & 7.331 \cellcolor{white} & 0.804 \cellcolor{white} & 2.073 \cellcolor{white} & 0.666 \cellcolor{white} & 4.001 \cellcolor{cnnresultcolor3} & 4.079 \cellcolor{cnnresultcolor3} & 0.923 \cellcolor{white} & 7.217 \cellcolor{white} & 122.4 \cellcolor{white} & 48.79 \cellcolor{white} & 97.38 \cellcolor{white} & 43.09 \cellcolor{white} & 85.88 \cellcolor{white} & 40.01 \cellcolor{white} & 80.27 \cellcolor{white} & 40.90 \cellcolor{white} & 81.98 \cellcolor{white} \\
			\hline
			im2col-copy-long-atb-ik & 20.51 \cellcolor{white} & 7.082 \cellcolor{white} & 10.64 \cellcolor{white} & 7.256 \cellcolor{white} & 0.806 \cellcolor{white} & 2.101 \cellcolor{white} & 0.663 \cellcolor{white} & 4.186 \cellcolor{white} & 4.238 \cellcolor{white} & 0.913 \cellcolor{white} & 7.733 \cellcolor{white} & 122.4 \cellcolor{white} & 48.90 \cellcolor{white} & 98.06 \cellcolor{white} & 42.24 \cellcolor{white} & 84.26 \cellcolor{white} & 40.23 \cellcolor{white} & 80.75 \cellcolor{white} & 42.32 \cellcolor{white} & 84.26 \cellcolor{white} \\
			\hline
			im2col-copy-long-atb-ki & 21.22 \cellcolor{white} & 7.316 \cellcolor{white} & 11.11 \cellcolor{white} & 7.484 \cellcolor{white} & 0.797 \cellcolor{white} & 2.091 \cellcolor{white} & 0.654 \cellcolor{cnnresultcolor3} & 4.054 \cellcolor{white} & 4.189 \cellcolor{white} & 0.980 \cellcolor{white} & 7.209 \cellcolor{white} & 122.1 \cellcolor{white} & 48.10 \cellcolor{white} & 97.42 \cellcolor{white} & 42.53 \cellcolor{white} & 85.85 \cellcolor{white} & 40.52 \cellcolor{white} & 81.10 \cellcolor{white} & 42.98 \cellcolor{white} & 86.10 \cellcolor{white} \\
			\hline
			im2col-copy-long-atbt-ik & 20.36 \cellcolor{white} & 7.007 \cellcolor{white} & 10.53 \cellcolor{white} & 7.198 \cellcolor{white} & 0.794 \cellcolor{white} & 2.091 \cellcolor{white} & 0.701 \cellcolor{white} & 4.193 \cellcolor{white} & 4.217 \cellcolor{white} & 0.886 \cellcolor{cnnresultcolor3} & 7.760 \cellcolor{white} & 122.4 \cellcolor{white} & 48.67 \cellcolor{white} & 98.33 \cellcolor{white} & 42.59 \cellcolor{white} & 84.37 \cellcolor{white} & 39.92 \cellcolor{white} & 79.84 \cellcolor{white} & 41.01 \cellcolor{white} & 83.48 \cellcolor{white} \\
			\hline
			im2col-copy-self-ab-ki & 21.11 \cellcolor{white} & 7.105 \cellcolor{white} & 10.72 \cellcolor{white} & 7.313 \cellcolor{white} & 0.924 \cellcolor{white} & 2.155 \cellcolor{white} & 0.666 \cellcolor{white} & 4.051 \cellcolor{white} & 4.097 \cellcolor{white} & 0.925 \cellcolor{white} & 7.201 \cellcolor{white} & 123.2 \cellcolor{white} & 48.78 \cellcolor{white} & 98.20 \cellcolor{white} & 43.09 \cellcolor{white} & 86.09 \cellcolor{white} & 39.97 \cellcolor{white} & 79.76 \cellcolor{white} & 40.95 \cellcolor{white} & 82.06 \cellcolor{white} \\
			\hline
			im2col-copy-self-atb-ik & 20.41 \cellcolor{white} & 7.059 \cellcolor{white} & 10.70 \cellcolor{white} & 7.275 \cellcolor{white} & 0.904 \cellcolor{white} & 2.154 \cellcolor{white} & 0.698 \cellcolor{white} & 4.216 \cellcolor{white} & 4.228 \cellcolor{white} & 0.906 \cellcolor{white} & 7.719 \cellcolor{white} & 123.1 \cellcolor{white} & 48.87 \cellcolor{white} & 98.48 \cellcolor{white} & 42.35 \cellcolor{white} & 85.13 \cellcolor{white} & 40.45 \cellcolor{white} & 80.57 \cellcolor{white} & 42.11 \cellcolor{white} & 84.58 \cellcolor{white} \\
			\hline
			im2col-copy-self-atb-ki & 21.25 \cellcolor{white} & 7.314 \cellcolor{white} & 11.15 \cellcolor{white} & 7.502 \cellcolor{white} & 0.911 \cellcolor{white} & 2.151 \cellcolor{white} & 0.659 \cellcolor{white} & 4.089 \cellcolor{white} & 4.223 \cellcolor{white} & 0.946 \cellcolor{white} & 7.197 \cellcolor{cnnresultcolor2} & 123.1 \cellcolor{white} & 48.89 \cellcolor{white} & 97.80 \cellcolor{white} & 43.44 \cellcolor{white} & 86.15 \cellcolor{white} & 40.20 \cellcolor{white} & 81.54 \cellcolor{white} & 42.95 \cellcolor{white} & 86.13 \cellcolor{white} \\
			\hline
			im2col-copy-self-atbt-ik & 20.44 \cellcolor{white} & 7.005 \cellcolor{white} & 10.50 \cellcolor{white} & 7.224 \cellcolor{white} & 0.892 \cellcolor{white} & 2.154 \cellcolor{white} & 0.682 \cellcolor{white} & 4.227 \cellcolor{white} & 4.211 \cellcolor{white} & 0.913 \cellcolor{white} & 7.756 \cellcolor{white} & 123.2 \cellcolor{white} & 48.77 \cellcolor{white} & 98.39 \cellcolor{white} & 42.68 \cellcolor{white} & 85.17 \cellcolor{white} & 40.05 \cellcolor{white} & 79.94 \cellcolor{white} & 41.07 \cellcolor{white} & 83.84 \cellcolor{white} \\
			\hline
			im2col-copy-short-ab-ki & 21.32 \cellcolor{white} & 7.116 \cellcolor{white} & 10.69 \cellcolor{white} & 7.298 \cellcolor{white} & 0.857 \cellcolor{white} & 2.122 \cellcolor{white} & 0.686 \cellcolor{white} & 4.033 \cellcolor{white} & 4.080 \cellcolor{white} & 0.900 \cellcolor{white} & 7.212 \cellcolor{white} & 123.4 \cellcolor{white} & 48.22 \cellcolor{white} & 97.78 \cellcolor{white} & 43.01 \cellcolor{white} & 84.96 \cellcolor{white} & 40.11 \cellcolor{white} & 80.23 \cellcolor{white} & 40.84 \cellcolor{white} & 81.89 \cellcolor{white} \\
			\hline
			im2col-copy-short-atb-ik & 20.59 \cellcolor{white} & 7.098 \cellcolor{white} & 10.63 \cellcolor{white} & 7.246 \cellcolor{white} & 0.821 \cellcolor{white} & 2.141 \cellcolor{white} & 0.699 \cellcolor{white} & 4.130 \cellcolor{white} & 4.220 \cellcolor{white} & 0.885 \cellcolor{cnnresultcolor2} & 7.724 \cellcolor{white} & 122.9 \cellcolor{white} & 48.52 \cellcolor{white} & 98.28 \cellcolor{white} & 42.57 \cellcolor{white} & 85.07 \cellcolor{white} & 40.42 \cellcolor{white} & 80.25 \cellcolor{white} & 42.08 \cellcolor{white} & 84.24 \cellcolor{white} \\
			\hline
			im2col-copy-short-atb-ki & 21.53 \cellcolor{white} & 7.344 \cellcolor{white} & 11.08 \cellcolor{white} & 7.470 \cellcolor{white} & 0.797 \cellcolor{white} & 2.119 \cellcolor{white} & 0.733 \cellcolor{white} & 4.033 \cellcolor{white} & 4.193 \cellcolor{white} & 0.941 \cellcolor{white} & 7.226 \cellcolor{white} & 123.1 \cellcolor{white} & 48.12 \cellcolor{white} & 98.05 \cellcolor{white} & 43.50 \cellcolor{white} & 85.38 \cellcolor{white} & 40.48 \cellcolor{white} & 80.26 \cellcolor{white} & 42.90 \cellcolor{white} & 86.02 \cellcolor{white} \\
			\hline
			im2col-copy-short-atbt-ik & 20.63 \cellcolor{white} & 6.991 \cellcolor{cnnresultcolor3} & 10.49 \cellcolor{cnnresultcolor3} & 7.150 \cellcolor{white} & 0.831 \cellcolor{white} & 2.102 \cellcolor{white} & 0.709 \cellcolor{white} & 4.146 \cellcolor{white} & 4.186 \cellcolor{white} & 0.930 \cellcolor{white} & 7.775 \cellcolor{white} & 123.2 \cellcolor{white} & 48.68 \cellcolor{white} & 98.54 \cellcolor{white} & 42.73 \cellcolor{white} & 85.02 \cellcolor{white} & 39.84 \cellcolor{white} & 80.15 \cellcolor{white} & 40.92 \cellcolor{white} & 83.77 \cellcolor{white} \\
			\hline
			im2col-scan-ab-ki & 23.36 \cellcolor{white} & 8.480 \cellcolor{white} & 12.55 \cellcolor{white} & 9.147 \cellcolor{white} & 6.294 \cellcolor{white} & 7.523 \cellcolor{white} & 1.191 \cellcolor{white} & 6.236 \cellcolor{white} & 5.067 \cellcolor{white} & 2.039 \cellcolor{white} & 11.03 \cellcolor{white} & 218.0 \cellcolor{white} & 69.38 \cellcolor{white} & 142.5 \cellcolor{white} & 54.48 \cellcolor{white} & 108.7 \cellcolor{white} & 45.40 \cellcolor{white} & 92.34 \cellcolor{white} & 43.69 \cellcolor{white} & 87.81 \cellcolor{white} \\
			\hline
			im2col-scan-atb-ik & 22.81 \cellcolor{white} & 8.291 \cellcolor{white} & 12.56 \cellcolor{white} & 9.126 \cellcolor{white} & 6.266 \cellcolor{white} & 7.497 \cellcolor{white} & 1.191 \cellcolor{white} & 6.153 \cellcolor{white} & 5.042 \cellcolor{white} & 2.071 \cellcolor{white} & 11.54 \cellcolor{white} & 218.1 \cellcolor{white} & 69.08 \cellcolor{white} & 143.0 \cellcolor{white} & 54.37 \cellcolor{white} & 108.4 \cellcolor{white} & 45.71 \cellcolor{white} & 92.52 \cellcolor{white} & 44.90 \cellcolor{white} & 90.40 \cellcolor{white} \\
			\hline
			im2col-scan-atb-ki & 23.50 \cellcolor{white} & 8.561 \cellcolor{white} & 12.94 \cellcolor{white} & 9.381 \cellcolor{white} & 6.296 \cellcolor{white} & 7.510 \cellcolor{white} & 1.195 \cellcolor{white} & 6.257 \cellcolor{white} & 5.175 \cellcolor{white} & 2.125 \cellcolor{white} & 11.02 \cellcolor{white} & 216.9 \cellcolor{white} & 69.30 \cellcolor{white} & 142.8 \cellcolor{white} & 54.73 \cellcolor{white} & 109.2 \cellcolor{white} & 46.08 \cellcolor{white} & 93.12 \cellcolor{white} & 45.77 \cellcolor{white} & 91.86 \cellcolor{white} \\
			\hline
			im2col-scan-atbt-ik & 22.71 \cellcolor{white} & 8.264 \cellcolor{white} & 12.36 \cellcolor{white} & 9.057 \cellcolor{white} & 6.281 \cellcolor{white} & 7.509 \cellcolor{white} & 1.150 \cellcolor{white} & 6.236 \cellcolor{white} & 5.111 \cellcolor{white} & 2.047 \cellcolor{white} & 11.56 \cellcolor{white} & 217.3 \cellcolor{white} & 69.50 \cellcolor{white} & 143.6 \cellcolor{white} & 54.19 \cellcolor{white} & 107.5 \cellcolor{white} & 45.22 \cellcolor{white} & 92.00 \cellcolor{white} & 44.05 \cellcolor{white} & 89.80 \cellcolor{white} \\
			\hline
			im2row-copy-short-ab-ik & 19.71 \cellcolor{cnnresultcolor2} & 6.926 \cellcolor{cnnresultcolor2} & 10.54 \cellcolor{white} & 7.088 \cellcolor{cnnresultcolor3} & 0.759 \cellcolor{cnnresultcolor2} & 2.044 \cellcolor{cnnresultcolor3} & 0.635 \cellcolor{cnnresultcolor2} & 3.960 \cellcolor{cnnresultcolor1} & 4.034 \cellcolor{cnnresultcolor1} & 0.890 \cellcolor{white} & 8.021 \cellcolor{white} & 113.2 \cellcolor{white} & 46.80 \cellcolor{cnnresultcolor3} & 91.79 \cellcolor{cnnresultcolor2} & 41.30 \cellcolor{white} & 82.44 \cellcolor{white} & 39.52 \cellcolor{white} & 78.82 \cellcolor{white} & 42.08 \cellcolor{white} & 84.17 \cellcolor{white} \\
			\hline
			im2row-copy-short-abt-ik & 19.62 \cellcolor{cnnresultcolor1} & 6.924 \cellcolor{cnnresultcolor1} & 10.37 \cellcolor{cnnresultcolor1} & 7.074 \cellcolor{cnnresultcolor1} & 0.784 \cellcolor{white} & 2.049 \cellcolor{white} & 0.631 \cellcolor{cnnresultcolor1} & 3.987 \cellcolor{cnnresultcolor2} & 4.062 \cellcolor{cnnresultcolor2} & 0.887 \cellcolor{white} & 7.981 \cellcolor{white} & 113.1 \cellcolor{white} & 46.70 \cellcolor{cnnresultcolor2} & 91.95 \cellcolor{cnnresultcolor3} & 41.12 \cellcolor{cnnresultcolor3} & 82.40 \cellcolor{cnnresultcolor3} & 39.25 \cellcolor{cnnresultcolor3} & 78.92 \cellcolor{white} & 40.73 \cellcolor{cnnresultcolor3} & 83.54 \cellcolor{white} \\
			\hline
			im2row-copy-short-abt-ki & 20.23 \cellcolor{white} & 7.125 \cellcolor{white} & 10.69 \cellcolor{white} & 7.297 \cellcolor{white} & 0.839 \cellcolor{white} & 2.167 \cellcolor{white} & 0.681 \cellcolor{white} & 4.206 \cellcolor{white} & 4.185 \cellcolor{white} & 0.934 \cellcolor{white} & 8.292 \cellcolor{white} & 117.4 \cellcolor{white} & 47.24 \cellcolor{white} & 93.92 \cellcolor{white} & 41.94 \cellcolor{white} & 83.00 \cellcolor{white} & 39.25 \cellcolor{white} & 78.52 \cellcolor{cnnresultcolor3} & 40.68 \cellcolor{cnnresultcolor1} & 81.77 \cellcolor{cnnresultcolor3} \\
			\hline
			im2row-copy-short-atbt-ki & 20.34 \cellcolor{white} & 7.310 \cellcolor{white} & 11.10 \cellcolor{white} & 7.468 \cellcolor{white} & 0.821 \cellcolor{white} & 2.165 \cellcolor{white} & 0.729 \cellcolor{white} & 4.225 \cellcolor{white} & 4.281 \cellcolor{white} & 0.959 \cellcolor{white} & 7.834 \cellcolor{white} & 117.6 \cellcolor{white} & 47.71 \cellcolor{white} & 94.30 \cellcolor{white} & 42.03 \cellcolor{white} & 83.07 \cellcolor{white} & 39.94 \cellcolor{white} & 79.92 \cellcolor{white} & 42.73 \cellcolor{white} & 85.85 \cellcolor{white} \\
			\hline
			im2row-scan-ab-ik & 20.12 \cellcolor{white} & 7.123 \cellcolor{white} & 10.81 \cellcolor{white} & 7.408 \cellcolor{white} & 1.360 \cellcolor{white} & 2.593 \cellcolor{white} & 0.730 \cellcolor{white} & 4.349 \cellcolor{white} & 4.259 \cellcolor{white} & 0.885 \cellcolor{cnnresultcolor1} & 8.372 \cellcolor{white} & 158.1 \cellcolor{white} & 53.11 \cellcolor{white} & 109.2 \cellcolor{white} & 44.49 \cellcolor{white} & 89.48 \cellcolor{white} & 40.92 \cellcolor{white} & 82.64 \cellcolor{white} & 42.77 \cellcolor{white} & 85.94 \cellcolor{white} \\
			\hline
			im2row-scan-abt-ik & 20.14 \cellcolor{white} & 7.185 \cellcolor{white} & 10.73 \cellcolor{white} & 7.444 \cellcolor{white} & 1.237 \cellcolor{white} & 2.569 \cellcolor{white} & 0.683 \cellcolor{white} & 4.337 \cellcolor{white} & 4.219 \cellcolor{white} & 0.927 \cellcolor{white} & 8.301 \cellcolor{white} & 155.8 \cellcolor{white} & 52.47 \cellcolor{white} & 107.6 \cellcolor{white} & 43.84 \cellcolor{white} & 88.52 \cellcolor{white} & 40.57 \cellcolor{white} & 82.03 \cellcolor{white} & 41.71 \cellcolor{white} & 84.95 \cellcolor{white} \\
			\hline
			im2row-scan-abt-ki & 20.62 \cellcolor{white} & 7.410 \cellcolor{white} & 11.04 \cellcolor{white} & 7.663 \cellcolor{white} & 1.346 \cellcolor{white} & 2.624 \cellcolor{white} & 0.750 \cellcolor{white} & 4.577 \cellcolor{white} & 4.354 \cellcolor{white} & 0.901 \cellcolor{white} & 8.121 \cellcolor{white} & 160.0 \cellcolor{white} & 53.00 \cellcolor{white} & 109.6 \cellcolor{white} & 44.75 \cellcolor{white} & 90.13 \cellcolor{white} & 40.83 \cellcolor{white} & 82.36 \cellcolor{white} & 41.39 \cellcolor{white} & 83.66 \cellcolor{white} \\
			\hline
			im2row-scan-atbt-ki & 20.87 \cellcolor{white} & 7.645 \cellcolor{white} & 11.46 \cellcolor{white} & 7.862 \cellcolor{white} & 1.341 \cellcolor{white} & 2.644 \cellcolor{white} & 0.757 \cellcolor{white} & 4.614 \cellcolor{white} & 4.459 \cellcolor{white} & 0.984 \cellcolor{white} & 8.255 \cellcolor{white} & 160.4 \cellcolor{white} & 53.41 \cellcolor{white} & 110.0 \cellcolor{white} & 44.91 \cellcolor{white} & 90.06 \cellcolor{white} & 41.08 \cellcolor{white} & 82.80 \cellcolor{white} & 43.46 \cellcolor{white} & 87.61 \cellcolor{white} \\
			\hline
			kn2col & 32.48 \cellcolor{white} & 7.977 \cellcolor{white} & 11.77 \cellcolor{white} & 7.452 \cellcolor{white} & 1.104 \cellcolor{white} & 3.047 \cellcolor{white} & 2.561 \cellcolor{white} & 5.416 \cellcolor{white} & 4.978 \cellcolor{white} & 0.929 \cellcolor{white} & 272.7 \cellcolor{white} & 340.8 \cellcolor{white} & 117.8 \cellcolor{white} & 158.5 \cellcolor{white} & 63.20 \cellcolor{white} & 97.16 \cellcolor{white} & 48.79 \cellcolor{white} & 86.81 \cellcolor{white} & 46.35 \cellcolor{white} & 88.49 \cellcolor{white} \\
			\hline
			kn2col-as & 27.21 \cellcolor{white} & 7.577 \cellcolor{white} & 10.92 \cellcolor{white} & 7.317 \cellcolor{white} & 1.117 \cellcolor{white} & 2.935 \cellcolor{white} & 1.446 \cellcolor{white} & 5.495 \cellcolor{white} & 4.823 \cellcolor{white} & 0.890 \cellcolor{white} & 307.0 \cellcolor{white} & 384.7 \cellcolor{white} & 80.04 \cellcolor{white} & 127.3 \cellcolor{white} & 53.49 \cellcolor{white} & 93.87 \cellcolor{white} & 44.46 \cellcolor{white} & 83.00 \cellcolor{white} & 44.01 \cellcolor{white} & 85.62 \cellcolor{white} \\
			\hline
			kn2row & 26.70 \cellcolor{white} & 7.798 \cellcolor{white} & 11.30 \cellcolor{white} & 7.445 \cellcolor{white} & 0.857 \cellcolor{white} & 2.362 \cellcolor{white} & 1.234 \cellcolor{white} & 4.707 \cellcolor{white} & 4.846 \cellcolor{white} & 0.904 \cellcolor{white} & 60.87 \cellcolor{white} & 126.4 \cellcolor{white} & 63.09 \cellcolor{white} & 99.45 \cellcolor{white} & 48.49 \cellcolor{white} & 85.11 \cellcolor{white} & 43.37 \cellcolor{white} & 82.72 \cellcolor{white} & 44.30 \cellcolor{white} & 84.15 \cellcolor{white} \\
			\hline
			kn2row-aa-ab & 19.74 \cellcolor{cnnresultcolor3} & 6.995 \cellcolor{white} & 10.45 \cellcolor{cnnresultcolor2} & 7.086 \cellcolor{cnnresultcolor2} & 0.761 \cellcolor{cnnresultcolor3} & 1.902 \cellcolor{cnnresultcolor1} & 0.676 \cellcolor{white} & 4.029 \cellcolor{white} & 4.105 \cellcolor{white} & 0.903 \cellcolor{white} & 15.95 \cellcolor{white} & 113.8 \cellcolor{white} & 47.52 \cellcolor{white} & 94.76 \cellcolor{white} & 39.75 \cellcolor{cnnresultcolor2} & 81.55 \cellcolor{cnnresultcolor2} & 38.16 \cellcolor{cnnresultcolor1} & 76.82 \cellcolor{cnnresultcolor1} & 40.70 \cellcolor{cnnresultcolor2} & 80.28 \cellcolor{cnnresultcolor1} \\
			\hline
			kn2row-aa-abt & 19.87 \cellcolor{white} & 7.016 \cellcolor{white} & 10.51 \cellcolor{white} & 7.126 \cellcolor{white} & 0.752 \cellcolor{cnnresultcolor1} & 1.923 \cellcolor{cnnresultcolor2} & 0.738 \cellcolor{white} & 4.068 \cellcolor{white} & 4.120 \cellcolor{white} & 1.015 \cellcolor{white} & 16.45 \cellcolor{white} & 104.0 \cellcolor{cnnresultcolor2} & 45.65 \cellcolor{cnnresultcolor1} & 87.85 \cellcolor{cnnresultcolor1} & 39.60 \cellcolor{cnnresultcolor1} & 78.50 \cellcolor{cnnresultcolor1} & 38.28 \cellcolor{cnnresultcolor2} & 76.95 \cellcolor{cnnresultcolor2} & 40.81 \cellcolor{white} & 80.65 \cellcolor{cnnresultcolor2} \\
			\hline
			kn2row-as & 23.83 \cellcolor{white} & 7.860 \cellcolor{white} & 11.21 \cellcolor{white} & 7.586 \cellcolor{white} & 0.846 \cellcolor{white} & 2.131 \cellcolor{white} & 1.194 \cellcolor{white} & 4.758 \cellcolor{white} & 4.762 \cellcolor{white} & 0.901 \cellcolor{white} & 44.26 \cellcolor{white} & 143.5 \cellcolor{white} & 62.01 \cellcolor{white} & 107.4 \cellcolor{white} & 47.31 \cellcolor{white} & 90.48 \cellcolor{white} & 41.23 \cellcolor{white} & 80.11 \cellcolor{white} & 42.76 \cellcolor{white} & 82.20 \cellcolor{white} \\
			\hline
			mec-col & 32.00 \cellcolor{white} & 14.52 \cellcolor{white} & 21.74 \cellcolor{white} & 14.80 \cellcolor{white} & 1.164 \cellcolor{white} & 2.667 \cellcolor{white} & 0.890 \cellcolor{white} & 6.093 \cellcolor{white} & 8.397 \cellcolor{white} & 2.474 \cellcolor{white} & 6.852 \cellcolor{cnnresultcolor1} & 101.9 \cellcolor{cnnresultcolor1} & 48.75 \cellcolor{white} & 96.43 \cellcolor{white} & 50.64 \cellcolor{white} & 101.5 \cellcolor{white} & 59.87 \cellcolor{white} & 124.4 \cellcolor{white} & 90.44 \cellcolor{white} & 199.5 \cellcolor{white} \\
			\hline
			mec-row-partition & 31.77 \cellcolor{white} & 14.51 \cellcolor{white} & 21.69 \cellcolor{white} & 14.60 \cellcolor{white} & 1.021 \cellcolor{white} & 2.482 \cellcolor{white} & 0.829 \cellcolor{white} & 5.929 \cellcolor{white} & 8.334 \cellcolor{white} & 2.516 \cellcolor{white} & 7.197 \cellcolor{cnnresultcolor3} & 107.0 \cellcolor{cnnresultcolor3} & 48.73 \cellcolor{white} & 97.28 \cellcolor{white} & 50.40 \cellcolor{white} & 101.7 \cellcolor{white} & 59.90 \cellcolor{white} & 125.1 \cellcolor{white} & 91.01 \cellcolor{white} & 199.4 \cellcolor{white} \\
			\hline
		\end{tabular}
}

\subfloat[multi-threaded]{
		\begin{tabular}{ |c|c|c|c|c|c|c|c|c|c|c|c|c|c|c|c|c|c|c|c|c|c| }
			\hline
			im2col-copy-long-ab-ki & 7.946 \cellcolor{white} & 3.230 \cellcolor{white} & 5.094 \cellcolor{white} & 3.348 \cellcolor{white} & 0.738 \cellcolor{white} & 0.667 \cellcolor{white} & 0.554 \cellcolor{white} & 1.447 \cellcolor{cnnresultcolor3} & 1.716 \cellcolor{white} & 1.296 \cellcolor{white} & 4.282 \cellcolor{cnnresultcolor3} & 64.99 \cellcolor{white} & 24.12 \cellcolor{white} & 61.82 \cellcolor{white} & 15.58 \cellcolor{white} & 29.74 \cellcolor{cnnresultcolor3} & 13.61 \cellcolor{white} & 27.02 \cellcolor{white} & 16.55 \cellcolor{white} & 33.51 \cellcolor{white} \\
			\hline
			im2col-copy-long-atb-ik & 7.267 \cellcolor{white} & 2.773 \cellcolor{white} & 4.060 \cellcolor{white} & 2.953 \cellcolor{white} & 0.427 \cellcolor{white} & 0.784 \cellcolor{white} & 0.408 \cellcolor{white} & 1.558 \cellcolor{white} & 1.458 \cellcolor{white} & 0.412 \cellcolor{white} & 9.306 \cellcolor{white} & 103.8 \cellcolor{white} & 26.89 \cellcolor{white} & 52.41 \cellcolor{white} & 15.48 \cellcolor{white} & 30.07 \cellcolor{white} & 12.83 \cellcolor{white} & 25.81 \cellcolor{white} & 13.55 \cellcolor{white} & 26.77 \cellcolor{white} \\
			\hline
			im2col-copy-long-atb-ki & 7.778 \cellcolor{white} & 3.285 \cellcolor{white} & 5.068 \cellcolor{white} & 3.401 \cellcolor{white} & 0.284 \cellcolor{white} & 0.671 \cellcolor{white} & 0.556 \cellcolor{white} & 1.445 \cellcolor{cnnresultcolor2} & 2.187 \cellcolor{white} & 0.904 \cellcolor{white} & 4.271 \cellcolor{cnnresultcolor1} & 65.07 \cellcolor{white} & 23.66 \cellcolor{white} & 61.72 \cellcolor{white} & 15.64 \cellcolor{white} & 29.80 \cellcolor{white} & 13.62 \cellcolor{white} & 27.12 \cellcolor{white} & 17.36 \cellcolor{white} & 34.92 \cellcolor{white} \\
			\hline
			im2col-copy-long-atbt-ik & 7.397 \cellcolor{white} & 2.740 \cellcolor{white} & 3.999 \cellcolor{white} & 2.924 \cellcolor{white} & 0.383 \cellcolor{white} & 0.801 \cellcolor{white} & 0.421 \cellcolor{white} & 1.469 \cellcolor{white} & 1.880 \cellcolor{white} & 0.627 \cellcolor{white} & 9.312 \cellcolor{white} & 104.2 \cellcolor{white} & 26.93 \cellcolor{white} & 52.33 \cellcolor{white} & 15.93 \cellcolor{white} & 30.21 \cellcolor{white} & 13.01 \cellcolor{white} & 25.80 \cellcolor{white} & 13.31 \cellcolor{white} & 27.09 \cellcolor{white} \\
			\hline
			im2col-copy-self-ab-ki & 7.788 \cellcolor{white} & 3.250 \cellcolor{white} & 4.897 \cellcolor{white} & 3.334 \cellcolor{white} & 0.478 \cellcolor{white} & 0.842 \cellcolor{white} & 0.582 \cellcolor{white} & 1.514 \cellcolor{white} & 1.722 \cellcolor{white} & 0.897 \cellcolor{white} & 4.310 \cellcolor{white} & 68.03 \cellcolor{white} & 24.60 \cellcolor{white} & 63.12 \cellcolor{white} & 16.11 \cellcolor{white} & 30.30 \cellcolor{white} & 13.60 \cellcolor{white} & 26.92 \cellcolor{white} & 16.59 \cellcolor{white} & 33.54 \cellcolor{white} \\
			\hline
			im2col-copy-self-atb-ik & 7.261 \cellcolor{white} & 2.754 \cellcolor{white} & 3.837 \cellcolor{cnnresultcolor1} & 2.968 \cellcolor{white} & 0.619 \cellcolor{white} & 0.964 \cellcolor{white} & 0.435 \cellcolor{white} & 1.557 \cellcolor{white} & 1.426 \cellcolor{white} & 0.409 \cellcolor{cnnresultcolor3} & 9.394 \cellcolor{white} & 106.5 \cellcolor{white} & 27.53 \cellcolor{white} & 53.82 \cellcolor{white} & 15.95 \cellcolor{white} & 30.53 \cellcolor{white} & 13.17 \cellcolor{white} & 26.00 \cellcolor{white} & 13.53 \cellcolor{white} & 26.81 \cellcolor{white} \\
			\hline
			im2col-copy-self-atb-ki & 7.846 \cellcolor{white} & 3.285 \cellcolor{white} & 6.020 \cellcolor{white} & 3.428 \cellcolor{white} & 0.474 \cellcolor{white} & 0.836 \cellcolor{white} & 0.591 \cellcolor{white} & 1.801 \cellcolor{white} & 1.969 \cellcolor{white} & 0.908 \cellcolor{white} & 4.347 \cellcolor{white} & 67.41 \cellcolor{white} & 24.34 \cellcolor{white} & 62.02 \cellcolor{white} & 16.27 \cellcolor{white} & 30.66 \cellcolor{white} & 14.05 \cellcolor{white} & 27.43 \cellcolor{white} & 17.38 \cellcolor{white} & 34.82 \cellcolor{white} \\
			\hline
			im2col-copy-self-atbt-ik & 7.244 \cellcolor{white} & 2.727 \cellcolor{cnnresultcolor3} & 3.851 \cellcolor{cnnresultcolor2} & 2.879 \cellcolor{cnnresultcolor3} & 0.577 \cellcolor{white} & 0.953 \cellcolor{white} & 0.436 \cellcolor{white} & 2.070 \cellcolor{white} & 1.440 \cellcolor{white} & 0.414 \cellcolor{white} & 9.389 \cellcolor{white} & 106.5 \cellcolor{white} & 27.76 \cellcolor{white} & 53.83 \cellcolor{white} & 16.37 \cellcolor{white} & 30.93 \cellcolor{white} & 13.27 \cellcolor{white} & 25.95 \cellcolor{white} & 13.29 \cellcolor{cnnresultcolor3} & 26.75 \cellcolor{cnnresultcolor3} \\
			\hline
			im2col-copy-short-ab-ki & 8.218 \cellcolor{white} & 3.202 \cellcolor{white} & 5.048 \cellcolor{white} & 3.336 \cellcolor{white} & 0.283 \cellcolor{white} & 0.666 \cellcolor{cnnresultcolor3} & 0.634 \cellcolor{white} & 1.948 \cellcolor{white} & 1.687 \cellcolor{white} & 0.886 \cellcolor{white} & 4.322 \cellcolor{white} & 66.15 \cellcolor{white} & 23.77 \cellcolor{white} & 62.12 \cellcolor{white} & 15.46 \cellcolor{cnnresultcolor3} & 29.84 \cellcolor{white} & 13.50 \cellcolor{white} & 27.50 \cellcolor{white} & 16.57 \cellcolor{white} & 33.89 \cellcolor{white} \\
			\hline
			im2col-copy-short-atb-ik & 7.479 \cellcolor{white} & 2.762 \cellcolor{white} & 4.057 \cellcolor{white} & 2.944 \cellcolor{white} & 0.387 \cellcolor{white} & 0.781 \cellcolor{white} & 0.511 \cellcolor{white} & 1.792 \cellcolor{white} & 1.713 \cellcolor{white} & 0.420 \cellcolor{white} & 9.323 \cellcolor{white} & 105.0 \cellcolor{white} & 26.96 \cellcolor{white} & 53.33 \cellcolor{white} & 15.89 \cellcolor{white} & 30.40 \cellcolor{white} & 12.93 \cellcolor{white} & 25.51 \cellcolor{white} & 13.47 \cellcolor{white} & 27.04 \cellcolor{white} \\
			\hline
			im2col-copy-short-atb-ki & 8.099 \cellcolor{white} & 3.279 \cellcolor{white} & 5.173 \cellcolor{white} & 3.359 \cellcolor{white} & 0.282 \cellcolor{white} & 0.666 \cellcolor{white} & 0.878 \cellcolor{white} & 1.912 \cellcolor{white} & 2.160 \cellcolor{white} & 1.096 \cellcolor{white} & 4.300 \cellcolor{white} & 66.67 \cellcolor{white} & 23.91 \cellcolor{white} & 61.88 \cellcolor{white} & 15.82 \cellcolor{white} & 29.86 \cellcolor{white} & 13.68 \cellcolor{white} & 27.03 \cellcolor{white} & 17.48 \cellcolor{white} & 34.65 \cellcolor{white} \\
			\hline
			im2col-copy-short-atbt-ik & 7.581 \cellcolor{white} & 2.737 \cellcolor{white} & 3.943 \cellcolor{cnnresultcolor3} & 2.886 \cellcolor{white} & 0.417 \cellcolor{white} & 0.781 \cellcolor{white} & 0.490 \cellcolor{white} & 1.999 \cellcolor{white} & 1.410 \cellcolor{cnnresultcolor2} & 0.846 \cellcolor{white} & 9.389 \cellcolor{white} & 104.9 \cellcolor{white} & 26.75 \cellcolor{white} & 52.55 \cellcolor{white} & 15.66 \cellcolor{white} & 30.35 \cellcolor{white} & 12.94 \cellcolor{white} & 25.73 \cellcolor{white} & 13.24 \cellcolor{cnnresultcolor2} & 26.98 \cellcolor{white} \\
			\hline
			im2col-scan-ab-ki & 14.23 \cellcolor{white} & 6.784 \cellcolor{white} & 9.784 \cellcolor{white} & 8.531 \cellcolor{white} & 13.99 \cellcolor{white} & 14.44 \cellcolor{white} & 1.839 \cellcolor{white} & 7.133 \cellcolor{white} & 4.095 \cellcolor{white} & 3.707 \cellcolor{white} & 15.82 \cellcolor{white} & 273.3 \cellcolor{white} & 80.56 \cellcolor{white} & 161.4 \cellcolor{white} & 45.02 \cellcolor{white} & 92.92 \cellcolor{white} & 27.94 \cellcolor{white} & 56.92 \cellcolor{white} & 24.18 \cellcolor{white} & 48.67 \cellcolor{white} \\
			\hline
			im2col-scan-atb-ik & 13.72 \cellcolor{white} & 6.010 \cellcolor{white} & 8.722 \cellcolor{white} & 7.695 \cellcolor{white} & 14.14 \cellcolor{white} & 14.54 \cellcolor{white} & 2.002 \cellcolor{white} & 7.005 \cellcolor{white} & 3.811 \cellcolor{white} & 3.199 \cellcolor{white} & 20.86 \cellcolor{white} & 310.2 \cellcolor{white} & 83.30 \cellcolor{white} & 154.4 \cellcolor{white} & 45.35 \cellcolor{white} & 93.78 \cellcolor{white} & 27.27 \cellcolor{white} & 56.11 \cellcolor{white} & 21.05 \cellcolor{white} & 42.23 \cellcolor{white} \\
			\hline
			im2col-scan-atb-ki & 14.33 \cellcolor{white} & 6.768 \cellcolor{white} & 9.868 \cellcolor{white} & 8.371 \cellcolor{white} & 14.04 \cellcolor{white} & 14.46 \cellcolor{white} & 1.850 \cellcolor{white} & 7.142 \cellcolor{white} & 4.270 \cellcolor{white} & 4.038 \cellcolor{white} & 15.83 \cellcolor{white} & 272.4 \cellcolor{white} & 80.78 \cellcolor{white} & 163.2 \cellcolor{white} & 44.78 \cellcolor{white} & 93.83 \cellcolor{white} & 27.98 \cellcolor{white} & 57.47 \cellcolor{white} & 24.86 \cellcolor{white} & 50.06 \cellcolor{white} \\
			\hline
			im2col-scan-atbt-ik & 13.67 \cellcolor{white} & 5.976 \cellcolor{white} & 8.834 \cellcolor{white} & 7.760 \cellcolor{white} & 14.13 \cellcolor{white} & 14.57 \cellcolor{white} & 1.882 \cellcolor{white} & 7.297 \cellcolor{white} & 3.942 \cellcolor{white} & 3.176 \cellcolor{white} & 20.86 \cellcolor{white} & 309.9 \cellcolor{white} & 84.23 \cellcolor{white} & 154.6 \cellcolor{white} & 45.17 \cellcolor{white} & 94.23 \cellcolor{white} & 27.35 \cellcolor{white} & 56.12 \cellcolor{white} & 20.77 \cellcolor{white} & 42.13 \cellcolor{white} \\
			\hline
			im2row-copy-short-ab-ik & 6.994 \cellcolor{cnnresultcolor2} & 2.703 \cellcolor{cnnresultcolor2} & 3.979 \cellcolor{white} & 2.845 \cellcolor{cnnresultcolor2} & 0.378 \cellcolor{white} & 0.773 \cellcolor{white} & 0.866 \cellcolor{white} & 1.469 \cellcolor{white} & 1.422 \cellcolor{cnnresultcolor3} & 0.612 \cellcolor{white} & 10.16 \cellcolor{white} & 102.1 \cellcolor{white} & 26.03 \cellcolor{white} & 52.03 \cellcolor{white} & 15.76 \cellcolor{white} & 30.65 \cellcolor{white} & 12.70 \cellcolor{cnnresultcolor2} & 25.21 \cellcolor{cnnresultcolor3} & 13.48 \cellcolor{white} & 26.72 \cellcolor{cnnresultcolor2} \\
			\hline
			im2row-copy-short-abt-ik & 6.970 \cellcolor{cnnresultcolor1} & 2.635 \cellcolor{cnnresultcolor1} & 3.949 \cellcolor{white} & 2.839 \cellcolor{cnnresultcolor1} & 0.492 \cellcolor{white} & 0.777 \cellcolor{white} & 0.647 \cellcolor{white} & 1.455 \cellcolor{white} & 1.374 \cellcolor{cnnresultcolor1} & 0.395 \cellcolor{cnnresultcolor1} & 10.26 \cellcolor{white} & 102.0 \cellcolor{white} & 26.15 \cellcolor{white} & 51.98 \cellcolor{white} & 15.59 \cellcolor{white} & 30.69 \cellcolor{white} & 12.72 \cellcolor{cnnresultcolor3} & 25.20 \cellcolor{cnnresultcolor2} & 12.99 \cellcolor{cnnresultcolor1} & 26.68 \cellcolor{cnnresultcolor1} \\
			\hline
			im2row-copy-short-abt-ki & 7.714 \cellcolor{white} & 3.246 \cellcolor{white} & 5.052 \cellcolor{white} & 3.339 \cellcolor{white} & 0.296 \cellcolor{white} & 0.676 \cellcolor{white} & 0.349 \cellcolor{cnnresultcolor3} & 1.423 \cellcolor{cnnresultcolor1} & 1.661 \cellcolor{white} & 1.274 \cellcolor{white} & 4.926 \cellcolor{white} & 66.85 \cellcolor{white} & 23.83 \cellcolor{white} & 60.39 \cellcolor{white} & 15.64 \cellcolor{white} & 30.67 \cellcolor{white} & 13.35 \cellcolor{white} & 26.32 \cellcolor{white} & 16.80 \cellcolor{white} & 33.51 \cellcolor{white} \\
			\hline
			im2row-copy-short-atbt-ki & 7.842 \cellcolor{white} & 3.241 \cellcolor{white} & 5.262 \cellcolor{white} & 3.395 \cellcolor{white} & 0.730 \cellcolor{white} & 0.972 \cellcolor{white} & 0.349 \cellcolor{white} & 1.453 \cellcolor{white} & 1.691 \cellcolor{white} & 0.905 \cellcolor{white} & 4.971 \cellcolor{white} & 68.23 \cellcolor{white} & 23.99 \cellcolor{white} & 61.82 \cellcolor{white} & 15.91 \cellcolor{white} & 30.94 \cellcolor{white} & 13.49 \cellcolor{white} & 26.93 \cellcolor{white} & 17.32 \cellcolor{white} & 34.98 \cellcolor{white} \\
			\hline
			im2row-scan-ab-ik & 7.674 \cellcolor{white} & 2.900 \cellcolor{white} & 4.400 \cellcolor{white} & 3.209 \cellcolor{white} & 0.963 \cellcolor{white} & 1.326 \cellcolor{white} & 0.571 \cellcolor{white} & 2.247 \cellcolor{white} & 1.744 \cellcolor{white} & 0.429 \cellcolor{white} & 10.44 \cellcolor{white} & 147.5 \cellcolor{white} & 33.81 \cellcolor{white} & 70.41 \cellcolor{white} & 19.77 \cellcolor{white} & 40.93 \cellcolor{white} & 15.01 \cellcolor{white} & 30.57 \cellcolor{white} & 14.32 \cellcolor{white} & 28.82 \cellcolor{white} \\
			\hline
			im2row-scan-abt-ik & 7.776 \cellcolor{white} & 3.078 \cellcolor{white} & 4.550 \cellcolor{white} & 3.403 \cellcolor{white} & 0.893 \cellcolor{white} & 1.315 \cellcolor{white} & 0.944 \cellcolor{white} & 2.599 \cellcolor{white} & 2.071 \cellcolor{white} & 0.410 \cellcolor{white} & 10.26 \cellcolor{white} & 147.0 \cellcolor{white} & 34.16 \cellcolor{white} & 70.17 \cellcolor{white} & 19.10 \cellcolor{white} & 37.89 \cellcolor{white} & 14.63 \cellcolor{white} & 30.00 \cellcolor{white} & 14.06 \cellcolor{white} & 28.99 \cellcolor{white} \\
			\hline
			im2row-scan-abt-ki & 8.392 \cellcolor{white} & 3.665 \cellcolor{white} & 5.637 \cellcolor{white} & 3.873 \cellcolor{white} & 0.802 \cellcolor{white} & 1.457 \cellcolor{white} & 0.468 \cellcolor{white} & 2.230 \cellcolor{white} & 2.357 \cellcolor{white} & 0.904 \cellcolor{white} & 4.972 \cellcolor{white} & 111.0 \cellcolor{white} & 31.10 \cellcolor{white} & 75.24 \cellcolor{white} & 19.27 \cellcolor{white} & 37.98 \cellcolor{white} & 15.24 \cellcolor{white} & 30.56 \cellcolor{white} & 17.45 \cellcolor{white} & 35.65 \cellcolor{white} \\
			\hline
			im2row-scan-atbt-ki & 8.593 \cellcolor{white} & 3.897 \cellcolor{white} & 5.729 \cellcolor{white} & 4.190 \cellcolor{white} & 0.827 \cellcolor{white} & 1.353 \cellcolor{white} & 0.763 \cellcolor{white} & 2.333 \cellcolor{white} & 2.249 \cellcolor{white} & 0.965 \cellcolor{white} & 4.988 \cellcolor{white} & 112.2 \cellcolor{white} & 30.48 \cellcolor{white} & 79.78 \cellcolor{white} & 19.38 \cellcolor{white} & 37.80 \cellcolor{white} & 15.31 \cellcolor{white} & 30.96 \cellcolor{white} & 18.32 \cellcolor{white} & 37.15 \cellcolor{white} \\
			\hline
			kn2col & 24.06 \cellcolor{white} & 3.915 \cellcolor{white} & 6.040 \cellcolor{white} & 3.799 \cellcolor{white} & 0.737 \cellcolor{white} & 2.050 \cellcolor{white} & 2.438 \cellcolor{white} & 3.843 \cellcolor{white} & 2.837 \cellcolor{white} & 0.849 \cellcolor{white} & 281.5 \cellcolor{white} & 295.2 \cellcolor{white} & 118.5 \cellcolor{white} & 130.3 \cellcolor{white} & 40.30 \cellcolor{white} & 50.55 \cellcolor{white} & 26.57 \cellcolor{white} & 38.65 \cellcolor{white} & 21.79 \cellcolor{white} & 39.59 \cellcolor{white} \\
			\hline
			kn2col-as & 17.67 \cellcolor{white} & 3.535 \cellcolor{white} & 4.644 \cellcolor{white} & 3.169 \cellcolor{white} & 0.741 \cellcolor{white} & 1.938 \cellcolor{white} & 1.854 \cellcolor{white} & 3.595 \cellcolor{white} & 2.874 \cellcolor{white} & 0.408 \cellcolor{cnnresultcolor2} & 376.6 \cellcolor{white} & 453.2 \cellcolor{white} & 69.82 \cellcolor{white} & 121.9 \cellcolor{white} & 32.39 \cellcolor{white} & 47.41 \cellcolor{white} & 20.84 \cellcolor{white} & 32.93 \cellcolor{white} & 17.11 \cellcolor{white} & 30.50 \cellcolor{white} \\
			\hline
			kn2row & 16.39 \cellcolor{white} & 4.575 \cellcolor{white} & 6.248 \cellcolor{white} & 4.170 \cellcolor{white} & 0.341 \cellcolor{white} & 1.051 \cellcolor{white} & 1.027 \cellcolor{white} & 2.351 \cellcolor{white} & 2.800 \cellcolor{white} & 1.185 \cellcolor{white} & 67.78 \cellcolor{white} & 74.16 \cellcolor{white} & 36.82 \cellcolor{white} & 48.19 \cellcolor{white} & 23.97 \cellcolor{white} & 34.48 \cellcolor{white} & 19.80 \cellcolor{white} & 32.64 \cellcolor{white} & 23.70 \cellcolor{white} & 41.40 \cellcolor{white} \\
			\hline
			kn2row-aa-ab & 7.064 \cellcolor{cnnresultcolor3} & 3.185 \cellcolor{white} & 4.646 \cellcolor{white} & 3.418 \cellcolor{white} & 0.239 \cellcolor{cnnresultcolor1} & 0.666 \cellcolor{cnnresultcolor2} & 0.308 \cellcolor{cnnresultcolor1} & 1.572 \cellcolor{white} & 1.694 \cellcolor{white} & 0.873 \cellcolor{white} & 15.16 \cellcolor{white} & 37.85 \cellcolor{cnnresultcolor2} & 15.91 \cellcolor{cnnresultcolor2} & 31.53 \cellcolor{cnnresultcolor2} & 12.09 \cellcolor{cnnresultcolor2} & 25.10 \cellcolor{cnnresultcolor1} & 12.39 \cellcolor{cnnresultcolor1} & 24.78 \cellcolor{cnnresultcolor1} & 16.61 \cellcolor{white} & 32.82 \cellcolor{white} \\
			\hline
			kn2row-aa-abt & 7.345 \cellcolor{white} & 3.171 \cellcolor{white} & 4.613 \cellcolor{white} & 3.211 \cellcolor{white} & 0.250 \cellcolor{cnnresultcolor2} & 0.655 \cellcolor{cnnresultcolor1} & 0.327 \cellcolor{cnnresultcolor2} & 1.620 \cellcolor{white} & 1.985 \cellcolor{white} & 0.868 \cellcolor{white} & 15.01 \cellcolor{white} & 34.21 \cellcolor{cnnresultcolor1} & 15.38 \cellcolor{cnnresultcolor1} & 29.17 \cellcolor{cnnresultcolor1} & 12.02 \cellcolor{cnnresultcolor1} & 25.46 \cellcolor{cnnresultcolor2} & 12.88 \cellcolor{white} & 25.64 \cellcolor{white} & 16.59 \cellcolor{white} & 33.23 \cellcolor{white} \\
			\hline
			kn2row-as & 11.99 \cellcolor{white} & 4.241 \cellcolor{white} & 5.714 \cellcolor{white} & 4.123 \cellcolor{white} & 0.266 \cellcolor{cnnresultcolor3} & 0.731 \cellcolor{white} & 0.956 \cellcolor{white} & 2.426 \cellcolor{white} & 2.645 \cellcolor{white} & 1.079 \cellcolor{white} & 45.06 \cellcolor{white} & 71.02 \cellcolor{white} & 30.43 \cellcolor{white} & 47.81 \cellcolor{white} & 19.40 \cellcolor{white} & 35.45 \cellcolor{white} & 16.08 \cellcolor{white} & 28.77 \cellcolor{white} & 19.88 \cellcolor{white} & 36.64 \cellcolor{white} \\
			\hline
			mec-col & 32.11 \cellcolor{white} & 14.51 \cellcolor{white} & 21.78 \cellcolor{white} & 14.82 \cellcolor{white} & 1.144 \cellcolor{white} & 1.420 \cellcolor{white} & 1.675 \cellcolor{white} & 7.647 \cellcolor{white} & 8.416 \cellcolor{white} & 2.465 \cellcolor{white} & 5.394 \cellcolor{white} & 51.75 \cellcolor{cnnresultcolor3} & 22.80 \cellcolor{cnnresultcolor3} & 43.65 \cellcolor{cnnresultcolor3} & 20.50 \cellcolor{white} & 40.38 \cellcolor{white} & 60.38 \cellcolor{white} & 124.5 \cellcolor{white} & 90.77 \cellcolor{white} & 200.6 \cellcolor{white} \\
			\hline
			mec-row-partition & 31.87 \cellcolor{white} & 15.81 \cellcolor{white} & 22.39 \cellcolor{white} & 15.87 \cellcolor{white} & 1.903 \cellcolor{white} & 1.217 \cellcolor{white} & 1.608 \cellcolor{white} & 8.471 \cellcolor{white} & 10.27 \cellcolor{white} & 4.075 \cellcolor{white} & 7.132 \cellcolor{white} & 59.92 \cellcolor{white} & 24.04 \cellcolor{white} & 47.01 \cellcolor{white} & 20.81 \cellcolor{white} & 41.76 \cellcolor{white} & 59.20 \cellcolor{white} & 125.0 \cellcolor{white} & 91.03 \cellcolor{white} & 200.5 \cellcolor{white} \\
			\hline
		\end{tabular}
}
\caption {Execution times (ms) on an Intel i7. Red is fastest, orange is second fastest, yellow is third fastest.}
\end{table*}
}
}
}
\clearpage

{ \footnotesize
{ \centering
{ \setlength{\tabcolsep}{0.093cm}
\begin{table*}[t]
\subfloat[single-threaded]{
	\begin{tabular}{ |c|c|c|c|c|c|c|c|c|c|c|c|c|c|c|c|c|c|c| }
	\hline
	height   &27 &13 &13 &13 &57 &57 &28 &28 &14 &7  &56 &56 &28 &28 &14  &14   \\
	\hline
	width    &27 &13 &13 &13 &57 &57 &28 &28 &14 &7  &56 &56 &28 &28 &14  &14   \\
	\hline
	channels &96 &256&384&384&64 &64 &16 &96 &160&832&128&256&256&512&512 &1024   \\
	\hline
	k        &5  &3  &3  &3  &1  &1  &5  &3  &3  &1  &3  &3  &3  &3  &3   &3   \\
	\hline
	kernels  &256&384&384&256&64 &192&32 &128&320&384&256&256&512&512&1024&1024     \\
	\hline
	\hline
	im2col-copy-long-ab-ki & 87.43 \cellcolor{white} & 30.51 \cellcolor{white} & 47.21 \cellcolor{white} & 30.37 \cellcolor{white} & 3.815 \cellcolor{white} & 9.266 \cellcolor{white} & 2.280 \cellcolor{white} & 17.63 \cellcolor{white} & 17.57 \cellcolor{white} & 3.434 \cellcolor{white} & 170.7 \cellcolor{white} & 343.8 \cellcolor{white} & 173.1 \cellcolor{white} & 346.9 \cellcolor{cnnresultcolor3} & 177.9 \cellcolor{white} & 358.9 \cellcolor{white}  \\
	\hline
	im2col-copy-long-atb-ik & 94.03 \cellcolor{white} & 30.78 \cellcolor{white} & 46.98 \cellcolor{white} & 31.63 \cellcolor{white} & 3.796 \cellcolor{white} & 9.726 \cellcolor{white} & 2.407 \cellcolor{white} & 18.95 \cellcolor{white} & 18.20 \cellcolor{white} & 3.331 \cellcolor{white} & 186.2 \cellcolor{white} & 364.1 \cellcolor{white} & 176.3 \cellcolor{white} & 358.6 \cellcolor{white} & 174.6 \cellcolor{white} & 356.3 \cellcolor{white} \\
	\hline
	im2col-copy-long-atb-ki & 92.08 \cellcolor{white} & 33.97 \cellcolor{white} & 46.76 \cellcolor{white} & 34.04 \cellcolor{white} & 3.809 \cellcolor{white} & 9.247 \cellcolor{white} & 2.338 \cellcolor{white} & 18.04 \cellcolor{white} & 18.25 \cellcolor{white} & 3.601 \cellcolor{white} & 169.2 \cellcolor{cnnresultcolor1} & 340.1 \cellcolor{cnnresultcolor3} & 175.5 \cellcolor{white} & 350.1 \cellcolor{white} & 189.1 \cellcolor{white} & 380.8 \cellcolor{white}  \\
	\hline
	im2col-copy-long-atbt-ik & 91.87 \cellcolor{white} & 30.62 \cellcolor{white} & 46.42 \cellcolor{white} & 33.15 \cellcolor{white} & 3.736 \cellcolor{white} & 10.00 \cellcolor{white} & 2.437 \cellcolor{white} & 18.37 \cellcolor{white} & 17.56 \cellcolor{white} & 3.325 \cellcolor{white} & 183.2 \cellcolor{white} & 368.3 \cellcolor{white} & 181.4 \cellcolor{white} & 353.3 \cellcolor{white} & 176.0 \cellcolor{white} & 365.9 \cellcolor{white}  \\
	\hline
	im2col-copy-self-ab-ki & 94.94 \cellcolor{white} & 30.21 \cellcolor{white} & 45.41 \cellcolor{white} & 30.36 \cellcolor{white} & 4.027 \cellcolor{white} & 9.400 \cellcolor{white} & 2.374 \cellcolor{white} & 18.93 \cellcolor{white} & 18.31 \cellcolor{white} & 3.546 \cellcolor{white} & 178.7 \cellcolor{white} & 341.3 \cellcolor{white} & 173.1 \cellcolor{white} & 350.1 \cellcolor{white} & 176.6 \cellcolor{white} & 355.9 \cellcolor{white}  \\
	\hline
	im2col-copy-self-atb-ik & 94.23 \cellcolor{white} & 30.26 \cellcolor{white} & 46.59 \cellcolor{white} & 31.75 \cellcolor{white} & 3.922 \cellcolor{white} & 9.391 \cellcolor{white} & 2.777 \cellcolor{white} & 18.56 \cellcolor{white} & 18.35 \cellcolor{white} & 3.298 \cellcolor{white} & 184.7 \cellcolor{white} & 366.2 \cellcolor{white} & 177.6 \cellcolor{white} & 366.4 \cellcolor{white} & 174.2 \cellcolor{cnnresultcolor3} & 356.6 \cellcolor{white}  \\
	\hline
	im2col-copy-self-atb-ki & 100.3 \cellcolor{white} & 31.62 \cellcolor{white} & 47.41 \cellcolor{white} & 32.48 \cellcolor{white} & 3.971 \cellcolor{white} & 9.729 \cellcolor{white} & 2.345 \cellcolor{white} & 18.07 \cellcolor{white} & 18.28 \cellcolor{white} & 3.590 \cellcolor{white} & 172.0 \cellcolor{white} & 343.5 \cellcolor{white} & 175.4 \cellcolor{white} & 352.8 \cellcolor{white} & 187.4 \cellcolor{white} & 375.4 \cellcolor{white}  \\
	\hline
	im2col-copy-self-atbt-ik & 96.57 \cellcolor{white} & 29.96 \cellcolor{white} & 45.53 \cellcolor{white} & 31.32 \cellcolor{white} & 3.905 \cellcolor{white} & 9.561 \cellcolor{white} & 2.447 \cellcolor{white} & 18.85 \cellcolor{white} & 17.82 \cellcolor{white} & 3.363 \cellcolor{white} & 187.2 \cellcolor{white} & 371.2 \cellcolor{white} & 178.7 \cellcolor{white} & 355.3 \cellcolor{white} & 175.9 \cellcolor{white} & 359.6 \cellcolor{white}  \\
	\hline
	im2col-copy-short-ab-ki & 86.44 \cellcolor{white} & 30.04 \cellcolor{white} & 44.69 \cellcolor{cnnresultcolor3} & 31.23 \cellcolor{white} & 3.808 \cellcolor{white} & 9.369 \cellcolor{white} & 2.297 \cellcolor{white} & 17.45 \cellcolor{white} & 17.86 \cellcolor{white} & 3.836 \cellcolor{white} & 170.0 \cellcolor{cnnresultcolor3} & 337.2 \cellcolor{cnnresultcolor2} & 171.8 \cellcolor{cnnresultcolor2} & 345.3 \cellcolor{cnnresultcolor1} & 175.7 \cellcolor{white} & 359.8 \cellcolor{white}  \\
	\hline
	im2col-copy-short-atb-ik & 90.47 \cellcolor{white} & 30.63 \cellcolor{white} & 45.82 \cellcolor{white} & 31.80 \cellcolor{white} & 3.738 \cellcolor{white} & 9.282 \cellcolor{white} & 2.539 \cellcolor{white} & 18.50 \cellcolor{white} & 17.76 \cellcolor{white} & 3.375 \cellcolor{white} & 181.3 \cellcolor{white} & 369.2 \cellcolor{white} & 176.6 \cellcolor{white} & 355.9 \cellcolor{white} & 175.0 \cellcolor{white} & 352.7 \cellcolor{white}  \\
	\hline
	im2col-copy-short-atb-ki & 87.12 \cellcolor{white} & 31.10 \cellcolor{white} & 46.69 \cellcolor{white} & 31.70 \cellcolor{white} & 3.867 \cellcolor{white} & 9.224 \cellcolor{cnnresultcolor3} & 2.292 \cellcolor{white} & 19.12 \cellcolor{white} & 20.81 \cellcolor{white} & 3.702 \cellcolor{white} & 169.9 \cellcolor{cnnresultcolor2} & 342.9 \cellcolor{white} & 174.1 \cellcolor{white} & 350.3 \cellcolor{white} & 185.7 \cellcolor{white} & 379.9 \cellcolor{white}  \\
	\hline
	im2col-copy-short-atbt-ik & 90.79 \cellcolor{white} & 29.85 \cellcolor{cnnresultcolor3} & 45.13 \cellcolor{white} & 30.68 \cellcolor{white} & 4.023 \cellcolor{white} & 9.236 \cellcolor{white} & 2.449 \cellcolor{white} & 18.03 \cellcolor{white} & 17.37 \cellcolor{cnnresultcolor3} & 3.361 \cellcolor{white} & 184.1 \cellcolor{white} & 367.9 \cellcolor{white} & 174.9 \cellcolor{white} & 359.4 \cellcolor{white} & 175.8 \cellcolor{white} & 366.4 \cellcolor{white}  \\
	\hline
	im2col-scan-ab-ki & 105.3 \cellcolor{white} & 39.04 \cellcolor{white} & 59.40 \cellcolor{white} & 44.90 \cellcolor{white} & 32.41 \cellcolor{white} & 37.78 \cellcolor{white} & 4.563 \cellcolor{white} & 30.08 \cellcolor{white} & 23.95 \cellcolor{white} & 10.45 \cellcolor{white} & 236.6 \cellcolor{white} & 494.0 \cellcolor{white} & 205.9 \cellcolor{white} & 412.0 \cellcolor{white} & 196.2 \cellcolor{white} & 401.8 \cellcolor{white}  \\
	\hline
	im2col-scan-atb-ik & 109.9 \cellcolor{white} & 39.88 \cellcolor{white} & 58.62 \cellcolor{white} & 45.81 \cellcolor{white} & 32.82 \cellcolor{white} & 38.69 \cellcolor{white} & 4.832 \cellcolor{white} & 31.43 \cellcolor{white} & 23.08 \cellcolor{white} & 9.062 \cellcolor{white} & 253.1 \cellcolor{white} & 505.4 \cellcolor{white} & 210.9 \cellcolor{white} & 427.0 \cellcolor{white} & 193.0 \cellcolor{white} & 387.7 \cellcolor{white}  \\
	\hline
	im2col-scan-atb-ki & 106.1 \cellcolor{white} & 47.51 \cellcolor{white} & 61.98 \cellcolor{white} & 44.93 \cellcolor{white} & 32.62 \cellcolor{white} & 37.79 \cellcolor{white} & 4.707 \cellcolor{white} & 30.54 \cellcolor{white} & 24.27 \cellcolor{white} & 9.457 \cellcolor{white} & 238.0 \cellcolor{white} & 478.9 \cellcolor{white} & 210.1 \cellcolor{white} & 425.4 \cellcolor{white} & 205.6 \cellcolor{white} & 414.0 \cellcolor{white}  \\
	\hline
	im2col-scan-atbt-ik & 109.3 \cellcolor{white} & 39.73 \cellcolor{white} & 58.05 \cellcolor{white} & 44.17 \cellcolor{white} & 33.77 \cellcolor{white} & 37.48 \cellcolor{white} & 5.414 \cellcolor{white} & 30.79 \cellcolor{white} & 23.34 \cellcolor{white} & 9.258 \cellcolor{white} & 251.4 \cellcolor{white} & 516.4 \cellcolor{white} & 215.5 \cellcolor{white} & 424.0 \cellcolor{white} & 194.0 \cellcolor{white} & 398.0 \cellcolor{white}  \\
	\hline
	im2row-copy-short-ab-ik & 86.39 \cellcolor{cnnresultcolor3} & 29.70 \cellcolor{cnnresultcolor2} & 45.64 \cellcolor{white} & 29.67 \cellcolor{cnnresultcolor1} & 3.597 \cellcolor{white} & 9.104 \cellcolor{cnnresultcolor2} & 2.248 \cellcolor{cnnresultcolor3} & 17.23 \cellcolor{cnnresultcolor3} & 17.23 \cellcolor{cnnresultcolor2} & 3.381 \cellcolor{white} & 178.8 \cellcolor{white} & 352.9 \cellcolor{white} & 173.1 \cellcolor{cnnresultcolor3} & 348.4 \cellcolor{white} & 172.1 \cellcolor{cnnresultcolor1} & 346.4 \cellcolor{cnnresultcolor1}  \\
	\hline
	im2row-copy-short-abt-ik & 85.69 \cellcolor{cnnresultcolor2} & 29.09 \cellcolor{cnnresultcolor1} & 43.63 \cellcolor{cnnresultcolor1} & 29.86 \cellcolor{cnnresultcolor2} & 3.443 \cellcolor{cnnresultcolor2} & 9.041 \cellcolor{cnnresultcolor1} & 2.195 \cellcolor{cnnresultcolor1} & 18.67 \cellcolor{white} & 16.88 \cellcolor{cnnresultcolor1} & 3.311 \cellcolor{white} & 181.4 \cellcolor{white} & 359.0 \cellcolor{white} & 173.8 \cellcolor{white} & 356.5 \cellcolor{white} & 173.6 \cellcolor{cnnresultcolor2} & 350.4 \cellcolor{cnnresultcolor2}  \\
	\hline
	im2row-copy-short-abt-ki & 85.36 \cellcolor{cnnresultcolor1} & 30.11 \cellcolor{white} & 44.35 \cellcolor{cnnresultcolor2} & 33.19 \cellcolor{white} & 3.993 \cellcolor{white} & 9.369 \cellcolor{white} & 2.234 \cellcolor{cnnresultcolor2} & 17.67 \cellcolor{white} & 18.43 \cellcolor{white} & 3.339 \cellcolor{white} & 171.6 \cellcolor{white} & 340.7 \cellcolor{white} & 170.3 \cellcolor{cnnresultcolor1} & 345.9 \cellcolor{cnnresultcolor2} & 175.0 \cellcolor{white} & 352.2 \cellcolor{white}  \\
	\hline
	im2row-copy-short-atbt-ki & 86.54 \cellcolor{white} & 30.78 \cellcolor{white} & 47.38 \cellcolor{white} & 32.39 \cellcolor{white} & 3.994 \cellcolor{white} & 9.459 \cellcolor{white} & 2.334 \cellcolor{white} & 17.74 \cellcolor{white} & 18.38 \cellcolor{white} & 3.440 \cellcolor{white} & 172.0 \cellcolor{white} & 336.5 \cellcolor{cnnresultcolor1} & 174.0 \cellcolor{white} & 357.8 \cellcolor{white} & 185.9 \cellcolor{white} & 371.2 \cellcolor{white}  \\
	\hline
	im2row-scan-ab-ik & 91.53 \cellcolor{white} & 30.36 \cellcolor{white} & 45.54 \cellcolor{white} & 31.11 \cellcolor{white} & 5.414 \cellcolor{white} & 11.93 \cellcolor{white} & 3.802 \cellcolor{white} & 19.01 \cellcolor{white} & 17.94 \cellcolor{white} & 3.476 \cellcolor{white} & 189.9 \cellcolor{white} & 384.7 \cellcolor{white} & 179.0 \cellcolor{white} & 361.5 \cellcolor{white} & 174.5 \cellcolor{white} & 350.5 \cellcolor{cnnresultcolor3}  \\
	\hline
	im2row-scan-abt-ik & 91.89 \cellcolor{white} & 30.54 \cellcolor{white} & 45.41 \cellcolor{white} & 32.41 \cellcolor{white} & 4.924 \cellcolor{white} & 10.70 \cellcolor{white} & 3.097 \cellcolor{white} & 18.29 \cellcolor{white} & 17.81 \cellcolor{white} & 3.644 \cellcolor{white} & 189.3 \cellcolor{white} & 389.0 \cellcolor{white} & 176.8 \cellcolor{white} & 365.8 \cellcolor{white} & 175.8 \cellcolor{white} & 354.2 \cellcolor{white}  \\
	\hline
	im2row-scan-abt-ki & 90.93 \cellcolor{white} & 29.91 \cellcolor{white} & 46.11 \cellcolor{white} & 30.20 \cellcolor{white} & 5.376 \cellcolor{white} & 10.93 \cellcolor{white} & 3.122 \cellcolor{white} & 18.37 \cellcolor{white} & 17.38 \cellcolor{white} & 3.442 \cellcolor{white} & 181.5 \cellcolor{white} & 374.2 \cellcolor{white} & 175.8 \cellcolor{white} & 356.8 \cellcolor{white} & 176.0 \cellcolor{white} & 355.8 \cellcolor{white}  \\
	\hline
	im2row-scan-atbt-ki & 91.21 \cellcolor{white} & 31.24 \cellcolor{white} & 49.26 \cellcolor{white} & 33.37 \cellcolor{white} & 5.286 \cellcolor{white} & 10.97 \cellcolor{white} & 3.125 \cellcolor{white} & 18.74 \cellcolor{white} & 18.08 \cellcolor{white} & 3.733 \cellcolor{white} & 181.2 \cellcolor{white} & 374.6 \cellcolor{white} & 178.4 \cellcolor{white} & 366.2 \cellcolor{white} & 187.3 \cellcolor{white} & 374.1 \cellcolor{white}  \\
	\hline
	kn2col & 131.7 \cellcolor{white} & 32.33 \cellcolor{white} & 45.22 \cellcolor{white} & 30.00 \cellcolor{white} & 4.162 \cellcolor{white} & 13.38 \cellcolor{white} & 7.109 \cellcolor{white} & 23.47 \cellcolor{white} & 22.28 \cellcolor{white} & 3.248 \cellcolor{cnnresultcolor3} & 501.6 \cellcolor{white} & 672.3 \cellcolor{white} & 208.3 \cellcolor{white} & 368.7 \cellcolor{white} & 186.8 \cellcolor{white} & 356.8 \cellcolor{white}  \\
	\hline
	kn2col-as & 123.5 \cellcolor{white} & 34.35 \cellcolor{white} & 48.51 \cellcolor{white} & 32.85 \cellcolor{white} & 4.794 \cellcolor{white} & 14.73 \cellcolor{white} & 5.755 \cellcolor{white} & 21.75 \cellcolor{white} & 20.86 \cellcolor{white} & 3.212 \cellcolor{cnnresultcolor2} & 336.8 \cellcolor{white} & 521.7 \cellcolor{white} & 215.7 \cellcolor{white} & 393.4 \cellcolor{white} & 193.5 \cellcolor{white} & 364.2 \cellcolor{white}  \\
	\hline
	kn2row & 109.4 \cellcolor{white} & 34.99 \cellcolor{white} & 48.10 \cellcolor{white} & 32.56 \cellcolor{white} & 3.742 \cellcolor{white} & 10.59 \cellcolor{white} & 4.739 \cellcolor{white} & 20.75 \cellcolor{white} & 19.64 \cellcolor{white} & 3.207 \cellcolor{cnnresultcolor1} & 196.5 \cellcolor{white} & 358.8 \cellcolor{white} & 188.8 \cellcolor{white} & 367.8 \cellcolor{white} & 189.2 \cellcolor{white} & 363.3 \cellcolor{white}  \\
	\hline
	kn2row-aa-ab & 90.84 \cellcolor{white} & 30.76 \cellcolor{white} & 47.07 \cellcolor{white} & 34.30 \cellcolor{white} & 3.370 \cellcolor{cnnresultcolor1} & 9.231 \cellcolor{white} & 2.252 \cellcolor{white} & 16.07 \cellcolor{cnnresultcolor1} & 17.43 \cellcolor{white} & 3.731 \cellcolor{white} & 176.5 \cellcolor{white} & 362.1 \cellcolor{white} & 179.5 \cellcolor{white} & 353.7 \cellcolor{white} & 186.8 \cellcolor{white} & 356.2 \cellcolor{white}  \\
	\hline
	kn2row-aa-abt & 90.20 \cellcolor{white} & 31.14 \cellcolor{white} & 48.21 \cellcolor{white} & 29.95 \cellcolor{cnnresultcolor3} & 3.580 \cellcolor{white} & 9.425 \cellcolor{white} & 2.309 \cellcolor{white} & 16.52 \cellcolor{cnnresultcolor2} & 18.08 \cellcolor{white} & 3.998 \cellcolor{white} & 177.9 \cellcolor{white} & 355.3 \cellcolor{white} & 179.8 \cellcolor{white} & 352.9 \cellcolor{white} & 185.2 \cellcolor{white} & 354.9 \cellcolor{white}  \\
	\hline
	kn2row-as & 102.6 \cellcolor{white} & 32.92 \cellcolor{white} & 47.04 \cellcolor{white} & 32.32 \cellcolor{white} & 3.469 \cellcolor{cnnresultcolor3} & 9.226 \cellcolor{white} & 3.143 \cellcolor{white} & 18.28 \cellcolor{white} & 18.81 \cellcolor{white} & 3.321 \cellcolor{white} & 189.0 \cellcolor{white} & 359.7 \cellcolor{white} & 188.5 \cellcolor{white} & 362.0 \cellcolor{white} & 188.5 \cellcolor{white} & 361.1 \cellcolor{white}  \\
	\hline
	mec-col & 120.5 \cellcolor{white} & 49.11 \cellcolor{white} & 72.45 \cellcolor{white} & 47.76 \cellcolor{white} & 4.677 \cellcolor{white} & 11.34 \cellcolor{white} & 2.810 \cellcolor{white} & 20.22 \cellcolor{white} & 25.58 \cellcolor{white} & 7.009 \cellcolor{white} & 191.6 \cellcolor{white} & 399.9 \cellcolor{white} & 218.1 \cellcolor{white} & 448.9 \cellcolor{white} & 299.0 \cellcolor{white} & 653.1 \cellcolor{white}  \\
	\hline
	mec-row-partition & 118.1 \cellcolor{white} & 47.85 \cellcolor{white} & 80.21 \cellcolor{white} & 47.43 \cellcolor{white} & 4.336 \cellcolor{white} & 10.18 \cellcolor{white} & 2.300 \cellcolor{white} & 18.97 \cellcolor{white} & 25.54 \cellcolor{white} & 6.237 \cellcolor{white} & 194.1 \cellcolor{white} & 394.3 \cellcolor{white} & 218.6 \cellcolor{white} & 437.7 \cellcolor{white} & 298.8 \cellcolor{white} & 647.9 \cellcolor{white}  \\
	\hline
	\end{tabular}
}
\\
\subfloat[multi-threaded]{
	\begin{tabular}{ |c|c|c|c|c|c|c|c|c|c|c|c|c|c|c|c|c|c| }
	\hline
	im2col-copy-long-ab-ki & 29.79 \cellcolor{white} & 11.30 \cellcolor{white} & 17.08 \cellcolor{white} & 10.96 \cellcolor{white} & 1.220 \cellcolor{white} & 2.889 \cellcolor{white} & 0.852 \cellcolor{white} & 5.858 \cellcolor{white} & 6.322 \cellcolor{white} & 1.276 \cellcolor{white} & 54.42 \cellcolor{white} & 116.3 \cellcolor{white} & 57.96 \cellcolor{white} & 110.2 \cellcolor{white} & 55.36 \cellcolor{white} & 112.5 \cellcolor{white}  \\
	\hline
	im2col-copy-long-atb-ik & 30.27 \cellcolor{white} & 10.22 \cellcolor{white} & 15.29 \cellcolor{white} & 12.16 \cellcolor{white} & 1.322 \cellcolor{white} & 3.357 \cellcolor{white} & 1.052 \cellcolor{white} & 6.138 \cellcolor{white} & 6.118 \cellcolor{white} & 1.294 \cellcolor{white} & 58.90 \cellcolor{white} & 119.4 \cellcolor{white} & 60.25 \cellcolor{white} & 114.5 \cellcolor{white} & 58.07 \cellcolor{white} & 121.1 \cellcolor{white} \\
	\hline
	im2col-copy-long-atb-ki & 30.98 \cellcolor{white} & 10.46 \cellcolor{white} & 16.03 \cellcolor{white} & 11.27 \cellcolor{white} & 1.194 \cellcolor{white} & 2.947 \cellcolor{white} & 0.897 \cellcolor{white} & 6.229 \cellcolor{white} & 6.371 \cellcolor{white} & 1.314 \cellcolor{white} & 56.20 \cellcolor{white} & 114.6 \cellcolor{white} & 56.68 \cellcolor{white} & 110.6 \cellcolor{white} & 57.14 \cellcolor{white} & 118.2 \cellcolor{white}  \\
	\hline
	im2col-copy-long-atbt-ik & 33.05 \cellcolor{white} & 10.70 \cellcolor{white} & 15.40 \cellcolor{white} & 12.18 \cellcolor{white} & 1.342 \cellcolor{white} & 3.357 \cellcolor{white} & 1.161 \cellcolor{white} & 6.220 \cellcolor{white} & 6.070 \cellcolor{white} & 1.711 \cellcolor{white} & 59.75 \cellcolor{white} & 121.9 \cellcolor{white} & 59.02 \cellcolor{white} & 114.5 \cellcolor{white} & 56.52 \cellcolor{white} & 120.7 \cellcolor{white}  \\
	\hline
	im2col-copy-self-ab-ki & 31.25 \cellcolor{white} & 10.29 \cellcolor{white} & 15.86 \cellcolor{white} & 10.98 \cellcolor{white} & 1.397 \cellcolor{white} & 2.890 \cellcolor{white} & 0.915 \cellcolor{white} & 6.038 \cellcolor{white} & 5.902 \cellcolor{white} & 1.279 \cellcolor{white} & 59.18 \cellcolor{white} & 118.2 \cellcolor{white} & 57.64 \cellcolor{white} & 110.7 \cellcolor{white} & 53.42 \cellcolor{cnnresultcolor3} & 112.8 \cellcolor{white}  \\
	\hline
	im2col-copy-self-atb-ik & 30.68 \cellcolor{white} & 12.16 \cellcolor{white} & 15.97 \cellcolor{white} & 11.10 \cellcolor{white} & 1.340 \cellcolor{white} & 3.445 \cellcolor{white} & 1.209 \cellcolor{white} & 6.368 \cellcolor{white} & 6.337 \cellcolor{white} & 1.307 \cellcolor{white} & 60.94 \cellcolor{white} & 122.7 \cellcolor{white} & 58.63 \cellcolor{white} & 117.8 \cellcolor{white} & 57.67 \cellcolor{white} & 118.5 \cellcolor{white}  \\
	\hline
	im2col-copy-self-atb-ki & 31.25 \cellcolor{white} & 10.61 \cellcolor{white} & 16.07 \cellcolor{white} & 11.50 \cellcolor{white} & 1.373 \cellcolor{white} & 2.880 \cellcolor{white} & 0.915 \cellcolor{white} & 6.603 \cellcolor{white} & 6.148 \cellcolor{white} & 1.374 \cellcolor{white} & 56.59 \cellcolor{white} & 116.9 \cellcolor{white} & 58.80 \cellcolor{white} & 112.7 \cellcolor{white} & 55.63 \cellcolor{white} & 128.6 \cellcolor{white}  \\
	\hline
	im2col-copy-self-atbt-ik & 30.62 \cellcolor{white} & 10.17 \cellcolor{white} & 15.55 \cellcolor{white} & 11.18 \cellcolor{white} & 1.435 \cellcolor{white} & 3.573 \cellcolor{white} & 1.247 \cellcolor{white} & 6.337 \cellcolor{white} & 6.042 \cellcolor{white} & 1.277 \cellcolor{white} & 59.97 \cellcolor{white} & 124.3 \cellcolor{white} & 58.79 \cellcolor{white} & 117.1 \cellcolor{white} & 59.24 \cellcolor{white} & 116.8 \cellcolor{white}  \\
	\hline
	im2col-copy-short-ab-ki & 29.96 \cellcolor{white} & 9.855 \cellcolor{white} & 15.06 \cellcolor{white} & 12.03 \cellcolor{white} & 1.188 \cellcolor{white} & 2.830 \cellcolor{white} & 0.839 \cellcolor{cnnresultcolor3} & 5.592 \cellcolor{white} & 5.705 \cellcolor{cnnresultcolor3} & 1.292 \cellcolor{white} & 54.69 \cellcolor{white} & 118.4 \cellcolor{white} & 54.97 \cellcolor{cnnresultcolor1} & 110.5 \cellcolor{white} & 53.84 \cellcolor{white} & 113.9 \cellcolor{white}  \\
	\hline
	im2col-copy-short-atb-ik & 29.81 \cellcolor{white} & 10.00 \cellcolor{white} & 15.38 \cellcolor{white} & 10.76 \cellcolor{white} & 1.343 \cellcolor{white} & 3.349 \cellcolor{white} & 1.098 \cellcolor{white} & 6.767 \cellcolor{white} & 5.938 \cellcolor{white} & 1.314 \cellcolor{white} & 58.08 \cellcolor{white} & 119.1 \cellcolor{white} & 58.53 \cellcolor{white} & 114.8 \cellcolor{white} & 57.30 \cellcolor{white} & 145.8 \cellcolor{white}  \\
	\hline
	im2col-copy-short-atb-ki & 31.17 \cellcolor{white} & 10.20 \cellcolor{white} & 16.98 \cellcolor{white} & 10.92 \cellcolor{white} & 1.225 \cellcolor{white} & 2.795 \cellcolor{cnnresultcolor3} & 0.851 \cellcolor{white} & 5.528 \cellcolor{white} & 5.876 \cellcolor{white} & 1.352 \cellcolor{white} & 53.42 \cellcolor{cnnresultcolor2} & 113.7 \cellcolor{cnnresultcolor2} & 55.82 \cellcolor{white} & 110.6 \cellcolor{white} & 56.22 \cellcolor{white} & 116.3 \cellcolor{white}  \\
	\hline
	im2col-copy-short-atbt-ik & 29.65 \cellcolor{white} & 10.20 \cellcolor{white} & 14.75 \cellcolor{cnnresultcolor3} & 10.51 \cellcolor{white} & 1.326 \cellcolor{white} & 3.352 \cellcolor{white} & 1.538 \cellcolor{white} & 6.007 \cellcolor{white} & 5.819 \cellcolor{white} & 1.273 \cellcolor{white} & 57.47 \cellcolor{white} & 118.6 \cellcolor{white} & 59.23 \cellcolor{white} & 118.0 \cellcolor{white} & 57.09 \cellcolor{white} & 116.4 \cellcolor{white}  \\
	\hline
	im2col-scan-ab-ki & 49.49 \cellcolor{white} & 19.48 \cellcolor{white} & 29.75 \cellcolor{white} & 25.68 \cellcolor{white} & 29.71 \cellcolor{white} & 30.86 \cellcolor{white} & 3.212 \cellcolor{white} & 19.03 \cellcolor{white} & 11.31 \cellcolor{white} & 6.926 \cellcolor{white} & 120.9 \cellcolor{white} & 248.1 \cellcolor{white} & 90.13 \cellcolor{white} & 180.8 \cellcolor{white} & 72.53 \cellcolor{white} & 151.4 \cellcolor{white}  \\
	\hline
	im2col-scan-atb-ik & 48.29 \cellcolor{white} & 19.29 \cellcolor{white} & 28.54 \cellcolor{white} & 24.05 \cellcolor{white} & 29.57 \cellcolor{white} & 32.48 \cellcolor{white} & 3.764 \cellcolor{white} & 19.04 \cellcolor{white} & 11.39 \cellcolor{white} & 7.629 \cellcolor{white} & 124.6 \cellcolor{white} & 255.5 \cellcolor{white} & 92.44 \cellcolor{white} & 184.3 \cellcolor{white} & 77.37 \cellcolor{white} & 156.1 \cellcolor{white}  \\
	\hline
	im2col-scan-atb-ki & 49.56 \cellcolor{white} & 19.02 \cellcolor{white} & 30.32 \cellcolor{white} & 24.13 \cellcolor{white} & 29.59 \cellcolor{white} & 30.78 \cellcolor{white} & 3.252 \cellcolor{white} & 18.34 \cellcolor{white} & 11.48 \cellcolor{white} & 7.086 \cellcolor{white} & 121.8 \cellcolor{white} & 249.4 \cellcolor{white} & 89.46 \cellcolor{white} & 180.5 \cellcolor{white} & 76.32 \cellcolor{white} & 156.4 \cellcolor{white}  \\
	\hline
	im2col-scan-atbt-ik & 48.99 \cellcolor{white} & 18.51 \cellcolor{white} & 30.47 \cellcolor{white} & 23.67 \cellcolor{white} & 30.48 \cellcolor{white} & 31.68 \cellcolor{white} & 3.536 \cellcolor{white} & 18.51 \cellcolor{white} & 11.63 \cellcolor{white} & 7.020 \cellcolor{white} & 124.7 \cellcolor{white} & 256.0 \cellcolor{white} & 90.68 \cellcolor{white} & 187.6 \cellcolor{white} & 76.87 \cellcolor{white} & 158.9 \cellcolor{white}  \\
	\hline
	im2row-copy-short-ab-ik & 30.09 \cellcolor{white} & 9.770 \cellcolor{cnnresultcolor3} & 14.80 \cellcolor{white} & 10.38 \cellcolor{cnnresultcolor3} & 1.130 \cellcolor{white} & 3.285 \cellcolor{white} & 1.031 \cellcolor{white} & 5.592 \cellcolor{white} & 5.816 \cellcolor{white} & 1.270 \cellcolor{white} & 56.67 \cellcolor{white} & 120.6 \cellcolor{white} & 59.23 \cellcolor{white} & 114.4 \cellcolor{white} & 56.50 \cellcolor{white} & 115.2 \cellcolor{white}  \\
	\hline
	im2row-copy-short-abt-ik & 29.38 \cellcolor{white} & 10.40 \cellcolor{white} & 14.78 \cellcolor{white} & 10.64 \cellcolor{white} & 1.259 \cellcolor{white} & 3.305 \cellcolor{white} & 1.026 \cellcolor{white} & 5.775 \cellcolor{white} & 5.904 \cellcolor{white} & 1.223 \cellcolor{white} & 58.58 \cellcolor{white} & 117.2 \cellcolor{white} & 57.92 \cellcolor{white} & 114.2 \cellcolor{white} & 58.28 \cellcolor{white} & 115.6 \cellcolor{white}  \\
	\hline
	im2row-copy-short-abt-ki & 29.37 \cellcolor{white} & 11.04 \cellcolor{white} & 15.00 \cellcolor{white} & 11.83 \cellcolor{white} & 1.356 \cellcolor{white} & 2.938 \cellcolor{white} & 0.859 \cellcolor{white} & 5.518 \cellcolor{cnnresultcolor3} & 6.204 \cellcolor{white} & 1.231 \cellcolor{white} & 53.35 \cellcolor{cnnresultcolor1} & 114.3 \cellcolor{white} & 55.53 \cellcolor{cnnresultcolor3} & 109.0 \cellcolor{cnnresultcolor3} & 52.98 \cellcolor{cnnresultcolor2} & 110.2 \cellcolor{cnnresultcolor3}  \\
	\hline
	im2row-copy-short-atbt-ki & 28.18 \cellcolor{cnnresultcolor3} & 11.25 \cellcolor{white} & 16.84 \cellcolor{white} & 11.27 \cellcolor{white} & 1.262 \cellcolor{white} & 2.832 \cellcolor{white} & 0.848 \cellcolor{white} & 5.557 \cellcolor{white} & 6.073 \cellcolor{white} & 1.339 \cellcolor{white} & 53.76 \cellcolor{cnnresultcolor3} & 116.5 \cellcolor{white} & 58.11 \cellcolor{white} & 111.2 \cellcolor{white} & 57.72 \cellcolor{white} & 116.4 \cellcolor{white}  \\
	\hline
	im2row-scan-ab-ik & 35.67 \cellcolor{white} & 10.54 \cellcolor{white} & 15.98 \cellcolor{white} & 11.69 \cellcolor{white} & 16.39 \cellcolor{white} & 6.532 \cellcolor{white} & 3.082 \cellcolor{white} & 7.153 \cellcolor{white} & 6.347 \cellcolor{white} & 1.427 \cellcolor{white} & 71.81 \cellcolor{white} & 159.5 \cellcolor{white} & 75.56 \cellcolor{white} & 125.7 \cellcolor{white} & 62.03 \cellcolor{white} & 120.7 \cellcolor{white}  \\
	\hline
	im2row-scan-abt-ik & 33.67 \cellcolor{white} & 10.36 \cellcolor{white} & 16.23 \cellcolor{white} & 11.28 \cellcolor{white} & 2.894 \cellcolor{white} & 4.835 \cellcolor{white} & 1.732 \cellcolor{white} & 7.799 \cellcolor{white} & 6.125 \cellcolor{white} & 1.386 \cellcolor{white} & 69.77 \cellcolor{white} & 156.4 \cellcolor{white} & 65.23 \cellcolor{white} & 125.9 \cellcolor{white} & 58.78 \cellcolor{white} & 119.4 \cellcolor{white}  \\
	\hline
	im2row-scan-abt-ki & 34.30 \cellcolor{white} & 10.60 \cellcolor{white} & 17.46 \cellcolor{white} & 12.65 \cellcolor{white} & 3.642 \cellcolor{white} & 4.275 \cellcolor{white} & 1.554 \cellcolor{white} & 6.653 \cellcolor{white} & 7.185 \cellcolor{white} & 1.428 \cellcolor{white} & 67.70 \cellcolor{white} & 160.0 \cellcolor{white} & 62.30 \cellcolor{white} & 118.3 \cellcolor{white} & 56.88 \cellcolor{white} & 113.8 \cellcolor{white}  \\
	\hline
	im2row-scan-atbt-ki & 34.59 \cellcolor{white} & 10.81 \cellcolor{white} & 17.14 \cellcolor{white} & 11.89 \cellcolor{white} & 2.917 \cellcolor{white} & 4.378 \cellcolor{white} & 1.458 \cellcolor{white} & 6.771 \cellcolor{white} & 6.205 \cellcolor{white} & 1.502 \cellcolor{white} & 67.43 \cellcolor{white} & 149.0 \cellcolor{white} & 60.96 \cellcolor{white} & 121.7 \cellcolor{white} & 58.73 \cellcolor{white} & 120.2 \cellcolor{white}  \\
	\hline
	kn2col & 67.26 \cellcolor{white} & 13.86 \cellcolor{white} & 18.38 \cellcolor{white} & 13.40 \cellcolor{white} & 2.238 \cellcolor{white} & 7.810 \cellcolor{white} & 4.874 \cellcolor{white} & 12.70 \cellcolor{white} & 9.116 \cellcolor{white} & 1.204 \cellcolor{white} & 374.8 \cellcolor{white} & 430.7 \cellcolor{white} & 90.76 \cellcolor{white} & 141.6 \cellcolor{white} & 71.27 \cellcolor{white} & 128.0 \cellcolor{white}  \\
	\hline
	kn2col-as & 59.39 \cellcolor{white} & 13.66 \cellcolor{white} & 17.86 \cellcolor{white} & 11.40 \cellcolor{white} & 2.172 \cellcolor{white} & 7.628 \cellcolor{white} & 4.468 \cellcolor{white} & 11.12 \cellcolor{white} & 9.873 \cellcolor{white} & 1.194 \cellcolor{cnnresultcolor2} & 192.8 \cellcolor{white} & 252.2 \cellcolor{white} & 95.45 \cellcolor{white} & 145.1 \cellcolor{white} & 67.29 \cellcolor{white} & 123.8 \cellcolor{white}  \\
	\hline
	kn2row & 42.41 \cellcolor{white} & 12.03 \cellcolor{white} & 16.32 \cellcolor{white} & 12.44 \cellcolor{white} & 1.126 \cellcolor{white} & 3.820 \cellcolor{white} & 2.593 \cellcolor{white} & 7.561 \cellcolor{white} & 7.210 \cellcolor{white} & 1.219 \cellcolor{white} & 72.83 \cellcolor{white} & 128.6 \cellcolor{white} & 65.08 \cellcolor{white} & 115.2 \cellcolor{white} & 62.01 \cellcolor{white} & 117.3 \cellcolor{white}  \\
	\hline
	kn2row-aa-ab & 26.75 \cellcolor{cnnresultcolor2} & 8.722 \cellcolor{cnnresultcolor1} & 13.26 \cellcolor{cnnresultcolor1} & 9.605 \cellcolor{cnnresultcolor2} & 0.915 \cellcolor{cnnresultcolor1} & 2.694 \cellcolor{cnnresultcolor1} & 0.813 \cellcolor{cnnresultcolor2} & 4.625 \cellcolor{cnnresultcolor2} & 5.461 \cellcolor{cnnresultcolor2} & 1.186 \cellcolor{cnnresultcolor1} & 54.18 \cellcolor{white} & 113.8 \cellcolor{cnnresultcolor3} & 55.62 \cellcolor{white} & 106.3 \cellcolor{cnnresultcolor1} & 53.74 \cellcolor{white} & 105.9 \cellcolor{cnnresultcolor1}  \\
	\hline
	kn2row-aa-abt & 24.46 \cellcolor{cnnresultcolor1} & 8.732 \cellcolor{cnnresultcolor2} & 14.36 \cellcolor{cnnresultcolor2} & 9.269 \cellcolor{cnnresultcolor1} & 0.925 \cellcolor{cnnresultcolor2} & 6.723 \cellcolor{white} & 0.677 \cellcolor{cnnresultcolor1} & 4.619 \cellcolor{cnnresultcolor1} & 4.674 \cellcolor{cnnresultcolor1} & 1.265 \cellcolor{white} & 54.30 \cellcolor{white} & 113.1 \cellcolor{cnnresultcolor1} & 55.40 \cellcolor{cnnresultcolor2} & 107.2 \cellcolor{cnnresultcolor2} & 50.78 \cellcolor{cnnresultcolor1} & 107.3 \cellcolor{cnnresultcolor2}  \\
	\hline
	kn2row-as & 39.89 \cellcolor{white} & 10.80 \cellcolor{white} & 15.62 \cellcolor{white} & 10.67 \cellcolor{white} & 0.961 \cellcolor{cnnresultcolor3} & 2.782 \cellcolor{cnnresultcolor2} & 1.540 \cellcolor{white} & 5.997 \cellcolor{white} & 6.444 \cellcolor{white} & 1.198 \cellcolor{cnnresultcolor3} & 69.36 \cellcolor{white} & 126.6 \cellcolor{white} & 62.39 \cellcolor{white} & 113.7 \cellcolor{white} & 66.23 \cellcolor{white} & 112.4 \cellcolor{white}  \\
	\hline
	mec-col & 53.44 \cellcolor{white} & 27.77 \cellcolor{white} & 39.48 \cellcolor{white} & 25.19 \cellcolor{white} & 4.462 \cellcolor{white} & 5.047 \cellcolor{white} & 1.666 \cellcolor{white} & 9.275 \cellcolor{white} & 12.82 \cellcolor{white} & 6.954 \cellcolor{white} & 91.76 \cellcolor{white} & 185.0 \cellcolor{white} & 107.5 \cellcolor{white} & 231.3 \cellcolor{white} & 159.8 \cellcolor{white} & 309.6 \cellcolor{white}  \\
	\hline
	mec-row-partition & 55.16 \cellcolor{white} & 27.12 \cellcolor{white} & 39.68 \cellcolor{white} & 26.37 \cellcolor{white} & 4.131 \cellcolor{white} & 4.650 \cellcolor{white} & 1.304 \cellcolor{white} & 8.633 \cellcolor{white} & 12.93 \cellcolor{white} & 6.671 \cellcolor{white} & 88.95 \cellcolor{white} & 184.7 \cellcolor{white} & 108.1 \cellcolor{white} & 209.8 \cellcolor{white} & 158.4 \cellcolor{white} & 308.4 \cellcolor{white}  \\
	\hline
	\end{tabular}
}
\caption {Execution times (ms) on an ARM Cortex A57. Red is fastest, orange is second fastest, yellow is third fastest.}
\end{table*}
}
}
}
\clearpage